\documentclass[runningheads]{llncs}
\usepackage{graphicx}
\usepackage[pagebackref,breaklinks,colorlinks]{hyperref}

\usepackage{tikz}
\usepackage{comment}
\usepackage{amsmath,amssymb} 
\usepackage[capitalize]{cleveref}
\usepackage{graphics} 
\usepackage{epsfig} 
\usepackage{empheq}
\usepackage{bm}
\usepackage{color}
\usepackage{bbding} 
\usepackage{cite}
\usepackage{diagbox}
\usepackage[linesnumbered,ruled]{algorithm2e}
\usepackage{hyperref}
\usepackage{float}
\usepackage{booktabs}
\usepackage{multirow}
\usepackage{mathrsfs}
\usepackage[utf8]{inputenc}
\usepackage{graphicx}
\usepackage{pifont}
\usepackage{threeparttable}
\crefname{section}{Sec.}{Secs.}
\Crefname{section}{Section}{Sections}
\Crefname{table}{Table}{Tables}
\crefname{table}{Tab.}{Tabs.}

\newcounter{RNum}
\renewcommand{\theRNum}{\arabic{RNum}}
\newcommand{\Remark}{\noindent\textbf{Remark}~\refstepcounter{RNum}\textbf{\theRNum}: }
\newcommand{\xmark}{\text{\ding{55}}}
\newcommand{\fref}[1]{Fig.~\ref{#1}}
\newcommand{\sref}[1]{Section~\ref{#1}}
\newcommand{\tref}[1]{Table~\ref{#1}}
\newcommand{\appref}[1]{Appendix~\ref{#1}}

\newcommand{\myparagraph}[1]{\noindent\textbf{#1}~}
\newcommand{\etal}{\textit{et al.}~}
\newcommand{\ie}{\textit{i.e.}}
\newcommand{\eg}{\textit{e.g.}}
\newcommand{\etc}{\textit{etc.}}
\newcommand{\green}[1]{\textcolor[rgb]{ 0,  .69,  .314}{#1}}
\newcommand{\red}[1]{\textcolor[rgb]{1,0,0}{#1}}

\begin{document}
\pagestyle{headings}
\mainmatter
\def\ECCVSubNumber{4293}  

\title{AirDet: Few-Shot Detection without Fine-tuning for Autonomous Exploration} 

\titlerunning{AirDet}
%
\author{Bowen Li\inst{1,2} \and
Chen Wang\inst{1} \and
Pranay Reddy\inst{1,3} \and\\
Seungchan Kim\inst{1} \and
Sebastian Scherer\inst{1}}
\authorrunning{B. Li, C. Wang, et al.}
%
\institute{Robotics Institute, Carnegie Mellon University, USA \\
\email{chenwang@dr.com, \{bowenli2,seungch2,basti\}@andrew.cmu.edu}\\ \and
School of Mechanical Engineering, Tongji University, China\\ \and
Electronics and Communication Engineering, IIITDM Jabalpur, India\\
\email{2018033@iiitdmj.ac.in}}
\maketitle

\begin{abstract}
Few-shot object detection has attracted increasing attention and rapidly progressed in recent years. However, the requirement of an exhaustive offline fine-tuning stage in existing methods is time-consuming and significantly hinders their usage in online applications such as autonomous exploration of low-power robots.
We find that their major limitation is that the little but valuable information from a few support images is not fully exploited.
To solve this problem, we propose a brand new architecture, AirDet, and surprisingly find that, by learning \textit{class-agnostic relation} with the support images in all modules, including cross-scale object proposal network, shots aggregation module, and localization network, AirDet without fine-tuning achieves comparable or even better results than many fine-tuned methods, reaching up to \textbf{30-40\%} improvements.
We also present solid results of onboard tests on real-world exploration data from the DARPA Subterranean Challenge, which strongly validate the feasibility of AirDet in robotics.
To the best of our knowledge, AirDet is the first feasible few-shot detection method for autonomous exploration of low-power robots.
The code and pre-trained models are released at \url{https://github.com/Jaraxxus-Me/AirDet}.

\end{abstract}

\section{Introduction}
\label{sec:intro}

\begin{figure}[!t]
	\centering
	\includegraphics[width=1\columnwidth]{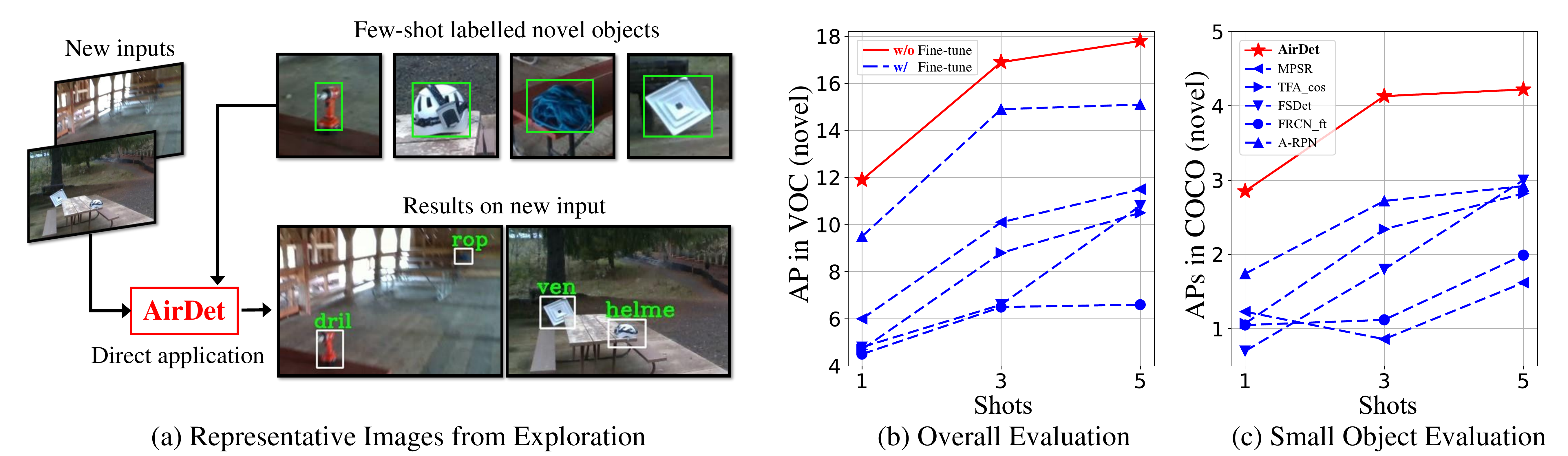}
	\caption{Representative images from robots’ exploration and performance comparison of the state-of-the-arts \cite{fan2020few,xiao2020few,wu2020multi,wang2020frustratingly,faster} and proposed AirDet. Solid lines denote results with no fine-tuning and dashed lines indicate results fine-tuned on few-shot data. Without further updating, AirDet can outperform prior work. Besides, unlike the fine-tuned models meeting bottleneck in small objects, AirDet has set an outstanding new level.}
	\label{fig:1}
\end{figure}

Few-shot object detection (FSOD) \cite{yan2019meta,wang2020frustratingly,xiao2020few,wu2020multi,zhu2021semantic} aims at detecting objects out of base training set with few support examples per class. 
It has received increasing attention from the robotics community due to its vital role in autonomous exploration, since robots are often expected to detect novel objects in an unknown environment but only a few examples can be provided online.
For example, in a mission of rescue shown in \fref{fig:1} (a), the robots are required to detect some uncommon objects such as drill, rope, helmet, vent.

Despite its recent promising developments, most of existing methods \cite{kang2019few,sun2021fsce,zhang2021accurate,wang2019meta,wang2020frustratingly,wu2020multi,qiao2021defrcn,cao2021nips,li2021few,fan2021generalized} require a careful \textit{offline} fine-tuning stage on novel images before inference. However, the requirement of the fine-tuning process is infeasible for robotic \textit{online} applications, since
(\textbf{1}) new object categories can be dynamically added during exploration, thus re-fine-tuning the model with limited onboard computational resources for novel classes is extremely inefficient for the time-starved tasks such as search and rescue \cite{tariq2018dronaid,farooq2018ground,wang2020visual,chen_tro}.
(\textbf{2}) to save human effort, only very few samples can be provided online\footnote{Since online annotation is needed during mission execution, only 1-5 samples can be provided in most of the robotic applications, which is the main focus of this paper.}, thus the fine-tuning stage \cite{kang2019few,sun2021fsce,zhang2021accurate,wang2019meta,wang2020frustratingly,wu2020multi,qiao2021defrcn,fan2021generalized,li2021few} needs careful \textit{offline} hyper-parameter tuning to avoid over-fitting, which is infeasible for \textit{online} exploration, and (\textbf{3}) fine-tuned models usually perform well for in-domain test \cite{wang2020frustratingly,wu2020multi,xiao2020few,faster,qiao2021defrcn,li2021few}, while suffer from cross-domain test, which is unfavourable for robotic applications.

Therefore, we often expect a few-shot detector that is able to inference without fine-tuning such as \cite{fan2020few}. However, the performance of \cite{fan2020few} is still severely hampered for challenging robotics domain due to (\textbf{1}) ineffective multi-scale detection;
(\textbf{2}) ineffective feature aggregation from multi-support images; and (\textbf{3}) inaccurate bounding box location prediction. Surprisingly, in this paper, we find that all three problems can be effectively solved by learning \textit{class-agnostic relation} with support images. We name the new architecture AirDet, which can produce promising results even without abominable fine-tuning as shown in \fref{fig:1}, which, to the best of our knowledge, is the first feasible few-shot detection model for autonomous robotic exploration. Specifically, the following three modules are proposed based on \textit{class-agnostic relation}, respectively.

\myparagraph{Support-guided Cross-Scale Fusion (SCS) for Object Proposal} One reason for performance degradation in multi-scale detection is that the region proposals are not effective for small scale objects, even though some existing works adopt multiple scale features from query images \cite{zhang2021accurate,zhu2021semantic}. We argue that the proposal network should also include cross-scale information from the support images. To this end, we present a novel SCS module, which explicitly extracts multi-scale features from cross-scale relations between support and query images.

\myparagraph{Global-Local Relation (GLR) for Shots Aggregation} Most prior work \cite{fan2020few,yan2019meta,wu2020multi} simply average multi-shot support feature to obtain a class prototype for detection head. However, this cannot fully exploit the little but valuable information from every support image. Instead, we construct a shots aggregation module by learning the relationship between the multi-support examples, which achieves significant improvements with more shots.

\myparagraph{Prototype Relation Embedding (PRE) for Location Regression} Some existing works \cite{fan2020few,zhang2021accurate} introduced a relation network \cite{sung2018learning} into the classification branch; however, the location regression branch is often neglected. To settle this, we introduce cross-correlation between the proposals from SCS and support features from GLR into the regression branch.
This results in the PRE module, which explicitly utilize support images for precise object localization.

In summary, AirDet is a fully relation-based few-shot object detector, which can be applied directly to the novel classes without fine-tuning. It surprisingly produces comparable or even better results than exhaustively fine-tuned SOTA methods \cite{wang2020frustratingly,faster,xiao2020few,wu2020multi,fan2020few}, as shown in \fref{fig:1} (b).
Besides, as shown in \fref{fig:1} (c), AirDet maintains high robustness in small objects due to the SCS module, which fully takes advantage of the multi-scale support feature.
Note that in this paper, fine-tuning is undesired because it cannot satisfy the online responsive requirement for robots, but it can still improve the performance of AirDet.






\section{Related Works}

\subsection{General Object Detection}

The task of object detection \cite{faster, yolo, liu2016ssd, RCNN,fast,mask} is to find out all the pre-defined objects in an image, predicting their categories and locations, which is one of the core problems in the field of computer vision.
Object detection algorithms are mainly divided into: two-stage approaches \cite{RCNN,fast,faster,mask} and one-stage approaches \cite{liu2016ssd,yolo,yolo2,yolo3}. R-CNN \cite{RCNN}, and its variants \cite{RCNN,fast,faster,mask} serve as the foundation of the former branch; among them, Faster R-CNN \cite{faster} used region proposal network (RPN) to generate class-agnostic proposals from the dense anchors, which greatly improved the speed of object detection based on R-CNN \cite{fast}. On the other hand, YOLO series \cite{yolo,yolo2,yolo3} fall into the second branch, which tackles object detection as an end-to-end regression problem. Besides, the known SSD series \cite{liu2016ssd,li2017fssd} propose to utilize pre-defined bounding boxes to adjust to various object scales inspired by \cite{faster}.

One shortcoming of the above methods is that they require abundant labeled data for training. Moreover, the types and number of object categories are fixed after training (80 classes in COCO, for instance), which is not applicable to robot's autonomous exploration, where unseen, novel objects often appear online.

\subsection{Few-shot Object Detection}
Trained with abundant data for base classes, few-shot object detectors can learn to generalize only using a few labeled novel image shots. Two main branches leading in FSOD are meta-learning-based approaches \cite{yan2019meta,xiao2020few,wu2020multi,fan2020few,han2021query} and transfer-learning-based approaches \cite{wang2020frustratingly,zhu2021semantic,sun2021fsce,wu2021universal,qiao2021defrcn}.

Transfer-learning approaches seek for the best learning strategy of general object detectors \cite{faster} on a few novel images. Wang \etal \cite{wang2020frustratingly} proposed to fine-tune only the last layer with a cosine similarity-based classifier. Using manually defined positive refinement branch, MPSR \cite{wu2020multi} mitigated the scale scarcity issue. Recent works have introduced semantic relations between novel-base classes \cite{zhu2021semantic} and contrastive proposal encoding \cite{sun2021fsce}.

Aiming at training meta-models on episodes of individual tasks, meta-learning approaches \cite{yan2019meta,xiao2020few,fan2020few,Hu2021CVPR,Zhang2021CVPR,han2021query} generally contain two branches, one for extracting support information and the other for detection on the query image. Among them, Meta R-CNN \cite{yan2019meta}, and FSDet \cite{xiao2020few} target at support guided query channel attention. With novel attention RPN and multi-relation classifier, A-RPN \cite{fan2020few} has set the current SOTA. Very recent works also cover support-query mutual guidance \cite{Zhang2021CVPR}, aggregating context information \cite{Hu2021CVPR}, and constructing heterogeneous graph convolutional networks on proposals \cite{han2021query}.

\subsection{Relation Network for Few-shot Learning}

In few-shot image classification, relation network \cite{sung2018learning}, also known as learning to compare, has been introduced to train a classifier by modeling the class-agnostic relation between a query image and the support images. Once trained and provided with a few novel support images, inference on novel query images can be implemented without further updating.

For few-shot object detection, such relation has only been utilized for the classification branch so far in very few works. For example, Fan \etal proposed a multi-relation classification network, which consists of global, local, and patch relation branches \cite{fan2020few}. Zhang \etal leveraged general relation network \cite{sung2018learning} architecture to build multi-level proposal scoring, and support weighting modules \cite{Zhang2021CVPR}. In this work, we thoroughly explore such relation in few-shot detection and propose a fully relation-based architecture.

\subsection{Multi-Scale Feature Extraction}
Multi-scale features have been exhaustively exploited for multi-scale objects in general object detection \cite{liu2016ssd,Shen2017dsod,Kong2016HyperNet,yolo2,lin2017fpn,li2017fssd}. For example, FSSD \cite{li2017fssd} proposed to fuse multi-scale feature and implement detection on the fused feature map. Lin \etal constructed the feature pyramid network (FPN) \cite{lin2017fpn}, which builds a top-down architecture and employs multi-scale feature map for detection.
For few-shot detection, standard FPN \cite{lin2017fpn} has been widely adopted in prior transfer-learning-based methods \cite{wang2020frustratingly,zhu2021semantic,sun2021fsce,wu2020multi}. In meta-learning, existing meta-learner \cite{Zhang2021CVPR} employs all scales from FPN and implements detection on each scale in parallel, which is computationally inefficient.

\section{Preliminary}
In few-shot object detection \cite{yan2019meta,xiao2020few,wu2020multi,Deng2009imagenet}, the classes are divided into $B$ base classes $\mathcal{C}_{\rm{b}}$ and $N$ novel ones $\mathcal{C}_{\rm{n}}$, satisfying that $\mathcal{C}_{\rm{b}}\cap\mathcal{C}_{\rm{n}}=\varnothing$. 
The objective is to train a model that can detect novel classes in $\mathcal{C}_{\rm{n}}$ by only providing $k$-shot labeled samples for $\mathcal{C}_{\rm{n}}$ and abundant images from base classes $\mathcal{C}_{\rm{b}}$.

During training, we adopt the episodic paradigm \cite{yan2019meta}.
Basically, images from the base classes $\mathcal{C}_{\rm{b}}$ are split into query images $\mathbf{Q}_{{\rm{b}}}$ and support images $\mathbf{S}_{{\rm{b}}}$.
Given all support images $\mathbf{S}_{{\rm{b}}}$, the model learns to detect objects in query images $\mathbf{Q}_{{\rm{b}}}$. During test, the model is to detect objects in novel query images $\mathbf{Q}_{{\rm{n}}}$ by only providing few (1-5) labeled novel support images $\mathbf{S}_{{\rm{n}}}$. 

\Remark Most existing methods \cite{wang2020frustratingly,zhu2021semantic,sun2021fsce,yan2019meta,xiao2020few,wu2020multi,wu2021universal,cao2021nips,qiao2021defrcn} have to be fine-tuned on $\mathbf{S}_{{\rm{n}}}$ due to the class-specific model design, while AirDet can be applied directly to $\mathbf{Q}_{{\rm{n}}}$ by providing $\mathbf{S}_{{\rm{n}}}$ without fine-tuning.

\section{Methodology}

\begin{figure*}[!t]
	\centering
	\includegraphics[width=1\textwidth]{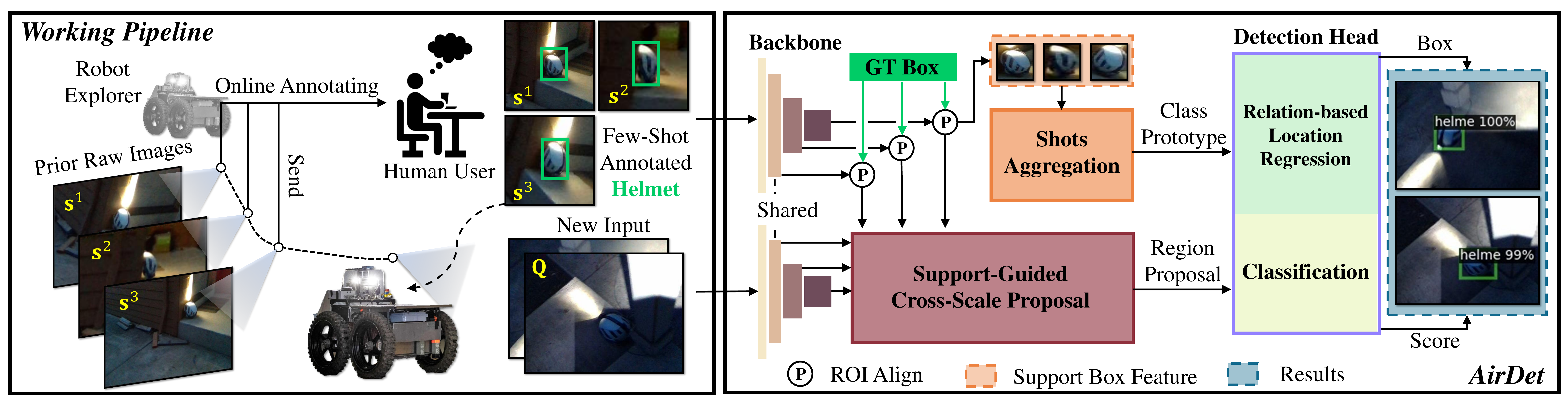}
	\caption{The pipeline of the autonomous exploration task and the framework of AirDet. During exploration, a few prior raw images that potentially contain novel objects (helmet) are sent to a human user first. Provided with online annotated few-shot data, the robot explorer is able to detect those objects by observing its surrounding environment. AirDet includes 4 modules, \ie, the shared backbone, support-guided cross-scale (SCS) feature fusion module for region proposal, global-local relation (GLR) module for shots aggregation, and relation-based detection head, which are visualized by different colors. }
	\label{fig:main}
\end{figure*}

Since only a few shots are given during model test, information from the support images is little but valuable.
We believe that the major limitation of the existing algorithms is that such information from support images is not fully exploited. 
Therefore, we propose to learn \textit{class-agnostic relation} with the support images in all the modules of AirDet.
As exhibited in \fref{fig:main}, the structure of AirDet is simple: except for the shared backbones, it only consists of three modules, \ie, a support-guided cross-scale fusion (SCS) module for regional proposal, a global-local relation (GLR) module for shots aggregation, and a relation-based detection head, containing prototype relation embedding (PRE) module for location regression and a multi-relation classifier \cite{fan2020few}. We next introduce two kinds of \textit{class-agnostic relation}, which will be used by the three modules.

\subsection{Class-Agnostic Relation}

To exploit the relation between two features from different aspects, we define two relation modules, \ie, spatial relation $\mathcal{R}_{\rm{s}}(\cdot, \cdot)$ and channel relation $\mathcal{R}_{\rm{c}}(\cdot, \cdot)$.

\myparagraph{1. Spatial Relation:}
Object features from the same category are often correlated along the spatial dimension, thus we define the spatial relation features $\mathcal{R}_{\rm{s}}$ in \eqref{eqn:inner} leveraging on the regular and depth-wise convolution.
\begin{equation}\label{eqn:inner}
    \mathcal{R}_{\rm{s}}(\mathbf{A}, \mathbf{B}) = \mathbf{A} \odot \mathrm{MLP}\Big(\mathrm{Flatten}\big(\mathrm{Conv}(\mathbf{B})\big)\Big),
\end{equation}
where inputs $\mathbf{A}, \mathbf{B}\in\mathbb{R}^{C\times W\times H}$ denote 2 general tensors. $\rm{Flatten}$ means flatten the features in spatial (image) domain and $\mathrm{MLP}$ denotes multilayer perceptron (MLP), so that $\rm{MLP}\Big(\rm{Flatten}\big(\rm{Conv}(\mathbf{B})\big)\Big) \in \mathbb{R}^{C\times 1\times 1}$. $\odot$ indicates depth-wise convolution \cite{fan2020few}. Note that we use convolution to calculate correlation since both operators are composed of inner products.

\myparagraph{2. Channel Relation:}
Inspired by the phenomenon that features of different classes are often stored in different channels \cite{li2019siamrpn}, we propose a simple but effective channel relation $\mathcal{R}_{\rm{c}}(\cdot, \cdot)$ in \eqref{eqn:channel} to extract the cross-class relation features.
\begin{equation}\label{eqn:channel}
    \mathcal{R}_{\rm{c}}(\mathbf{A}, \mathbf{B}) = \mathrm{{Conv}}\big(\mathrm{Cat}(\mathbf{A}, \mathbf{B})\big) + \mathrm{Cat}\big(\mathrm{{Conv}}(\mathbf{A}), \mathrm{{Conv}}(\mathbf{B})\big),
\end{equation}
where $\mathrm{Cat(\cdot, \cdot)}$ is to concatenate features along the channel dimension.

\Remark The two simple but effective \textit{class-agnostic relation} learners are fundamental building blocks of AirDet, which, to the best of our knowledge, is the first attempt towards a fully relation-based structure in few-shot detection.

\subsection{Support-guided Cross-Scale Fusion (SCS) for Object Proposal}

As mentioned earlier, existing works generate object proposals only using single scale information from query images \cite{kang2019few,xiao2020few,wang2019meta,wu2020multi}, while such strategy may not be effective for small scale novel objects.

\begin{figure}[ht]
	\centering
	\includegraphics[width=1\columnwidth]{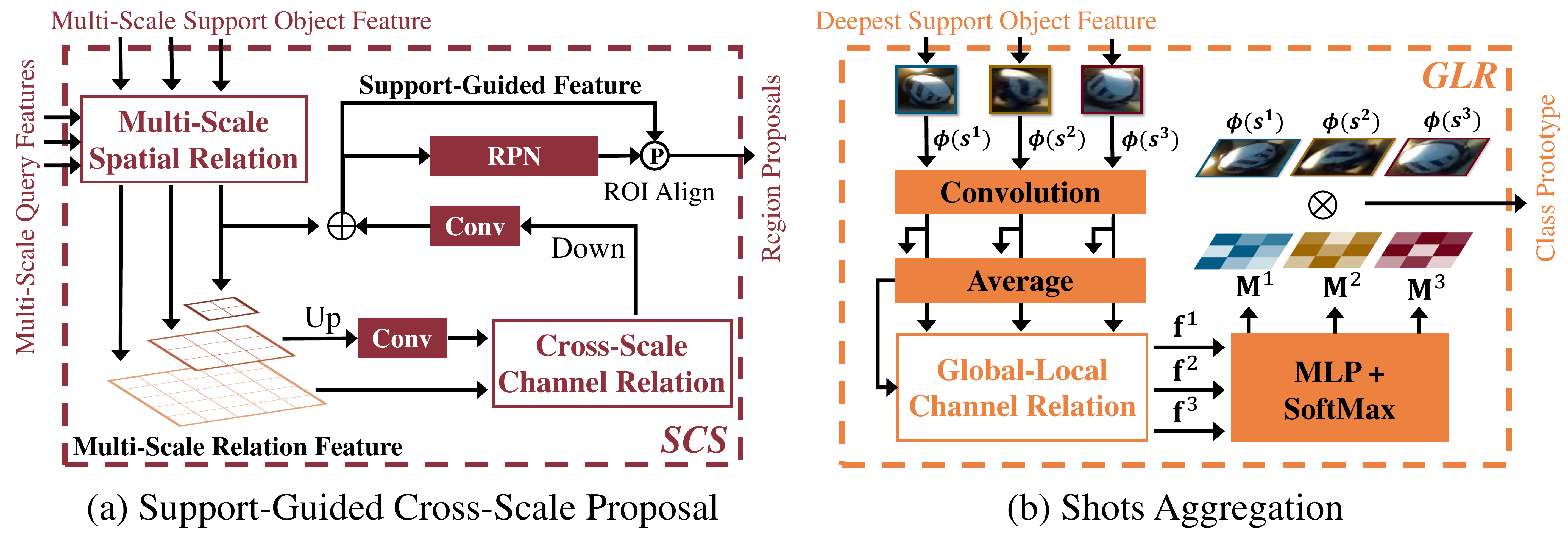}
	\caption{Network architecture of SCS for region proposal and GLR for shots aggregation.}
	\label{fig:scs}
\end{figure}

Differently, we propose support-guided cross-scale fusion (SCS) in AirDet to introduce multi-scale features and take the relation between query and support images for region proposal.
As shown in \fref{fig:scs} (a), SCS takes support and query features from different backbone blocks (ResNet2, 3, and 4 block) as input.
We first apply \textit{spatial relation}, where the query and support features from the same backbone block are $\mathbf{A}, \mathbf{B}$ in \eqref{eqn:inner}, respectively.
Then we use \textit{channel relation} to fuse the ResNet2 and ResNet3 block features, which are $\mathbf{A}, \mathbf{B}$ in \eqref{eqn:channel}, respectively. 
The fused channel relation feature is later merged with the spatial relation feature from ResNet4 block.
The final merged feature is sent to the region proposal network (RPN) \cite{faster} to generate region proposals.



\subsection{Global-Local Relation (GLR) for Shots Aggregation}

In prior attempts \cite{yan2019meta,kang2019few,xiao2020few,wu2020multi}, the support object features from multiple shots are usually averaged to represent the class prototype, which is then used for regression and classification.
Although it can be effective with fine-tuning, we argue that a simple average cannot fully utilize the information from few-shot data. To this end, we built global-local relation (GLR) module in \fref{fig:scs} (b), which leverages the features from every shot to construct the final prototype.

Suppose the $k$-shot deepest support features are $\phi({\mathbf{s}^i})$,~$i=1,\cdots,k$, our final class prototype $\mathbf{e}$ can be expressed as a weighted average of the features:
\begin{equation}
\mathbf{e}=\sum_{i=1}^{k}(\phi({\mathbf{s}^i})\otimes\mathbf{M}^i),
\end{equation}
where $\otimes$ is the element-wise multiplication, and $\mathbf{M}^i$ is a confidence map:
\begin{equation}
\mathbf{M}^{i}=\mathrm{SoftMax}\Big(\mathrm{MLP}\big(\mathbf{f}^{i}\big)\Big),
\end{equation}
where $\mathbf{f}^{i}$ is the output from the channel relation extractor:
\begin{equation}\label{eq:shot-relation}
    \mathbf{f}^{i} = \mathcal{R}_{\rm{c}}\left(\mathrm{Conv}(\phi(\mathbf{s}^i)), \frac{1}{k}\sum_{i=1}^{k}\mathrm{Conv}(\phi(\mathbf{s}^i))\right).
\end{equation}

Note that to include both ``global" (all shots) and ``local" (single shot) features, the inputs of the channel relation extractor in \eqref{eq:shot-relation} include both the feature from that shot and the averaged features over all shots.




\Remark Unlike prior work \cite{yan2019meta,kang2019few,xiao2020few,wu2020multi} relying on fine-tuning with more support data for performance gain, AirDet can extract stronger prototype to achieve improvement on more shots without fine-tuning.

\subsection{Prototype Relation Embedding (PRE) for Location Regression}
It has been demonstrated that a multi-relation network \cite{fan2020few} is effective for the classification branch.
Inspired by its success, we further build a prototype relation embedding (PRE) network for the location regression branch.
Given a prototype exemplar $\mathbf{e}\in\mathbb{R}^{C\times a\times a}$, we utilize the spatial relation \eqref{eqn:inner} to embed information from the exemplar to the proposal features $\mathbf{p}^j$,~$j= 1,2,\cdots,p$ as:
\begin{equation}
    \mathbf{l}^j = \mathbf{p}^j + \mathcal{R}_{\rm{s}}(\mathbf{p}^j, \mathbf{e}),
\end{equation}
where we take $3\times 3$ convolution layer in \eqref{eqn:inner} for spatial feature extraction.
The proposal features $\mathbf{l}^j$ is then used for bounding box regression through an MLP module following Faster-RCNN \cite{faster}.

\Remark Class-related feature $\mathbf{l}^j$ contains information from support objects, which turns out more effective for location regression even if the objects have never been seen in the training set.

\section{Experiments}

\subsection{Implementation}\label{sec:Imple}

We adopt the training pipeline from \cite{fan2020few}.
To maintain a fair comparison with other methods \cite{wang2020frustratingly,xiao2020few,wu2020multi,faster}, we mainly adopt ResNet101 \cite{He2016res} pre-trained on ImageNet \cite{Deng2009imagenet} as backbone.
The performance of other backbones is presented in \appref{sec:backbone}.
For fair comparison \cite{wang2020frustratingly,xiao2020few,wu2020multi,fan2020few,faster}, we utilized their official implementation, support examples, and models (if provided) in all the experiments. AirDet and the baseline \cite{fan2020few} take the \textbf{same} supports in all the settings.
We use 4 NVIDIA GeForce Titan-X Pascal GPUs for experiments. 
Detailed configuration of AirDet can be found in \appref{sec:config} and the source code.

\Remark To save human effort, only very few support examples (1-5 samples per class) can be provided during online exploration. Therefore, we mainly focused on $k=1, 2, 3, 5$-shot evaluation. Since the objects from exploration are usually unseen, we only test novel classes throughout the experiments.

\subsection{In-domain Evaluation}\label{sec:indomain}

We first present the in-domain evaluation on COCO benchmark \cite{lin2014microsoft}, where the models are trained and tested on the same dataset.
Following prior works \cite{yan2019meta,kang2019few,wu2020multi,fan2020few,wang2020frustratingly,xiao2020few,wu2021universal,cao2021nips,fan2021generalized,sun2021fsce,zhu2021semantic}, the 80 classes are split into 60 non-VOC base classes and 20 novel classes. During training, the base class images from COCO trainval2014 are considered available. With few-shot samples per novel class, the models are evaluated on 5,000 images from COCO val2014 dataset.

\begin{table}[!t]
	\centering
	\setlength{\tabcolsep}{0.2mm}
	\fontsize{5.5}{6.5}\selectfont
	\caption{Performance comparison on COCO validation dataset. In each setting, \red{red} and \green{green} fonts denote the best and second-best performance, respectively. AirDet achieves significant performance gain on baseline without fine-tuning. With fine-tuning, AirDet sets a new SOTA performance. $^\dag$We randomly sampled 3-5 different groups of support examples and reported the average performance and their standard deviation.}
	\begin{threeparttable}
	\begin{tabular}{cc|ccc|ccc|ccc|ccc}
		\toprule
		\multicolumn{2}{c|}{Shots} & \multicolumn{3}{c|}{1} & \multicolumn{3}{c|}{2} & \multicolumn{3}{c|}{3} & \multicolumn{3}{c}{5} \\
		\midrule
		Method  & Fine-tune & AP & AP$_{50}$ & AP$_{75}$ & AP & AP$_{50}$ & AP$_{75}$ & AP & AP$_{50}$ & AP$_{75}$ & AP & AP$_{50}$ & AP$_{75}$  \\
		
		\multirow{2}{*}{A-RPN \cite{fan2020few}}$\dag$ & \multirow{2}{*}{\xmark}    & 4.32  & 7.62 & 4.3 & 4.67  & 8.83  & 4.49  & 5.28  & 9.95 & 5.05 & 6.08  & 11.17 & 5.88 \\
		
          &   & $\pm$0.7 &$\pm$1.3  & $\pm$0.7 & $\pm$0.3  & $\pm$0.5  & $\pm$0.3  & $\pm$0.6  & $\pm$0.8  & $\pm$0.6 & $\pm$0.3  & $\pm$0.4 & $\pm$0.3 \\
        \cmidrule{1-14}
		\multirow{2}{*}{\textbf{AirDet (Ours)}}$\dag$  & \multirow{2}{*}{\xmark}     & \red{\textbf{5.97}} & \red{\textbf{10.52}} & \red{\textbf{5.98}} & \red{\textbf{6.58}} & \red{\textbf{12.02}} & \red{\textbf{6.33}} & \red{\textbf{7.00}} & \red{\textbf{12.95}} & \red{\textbf{6.71}} & \red{\textbf{7.76}} & \red{\textbf{14.28}} & \red{\textbf{
	7.31}} \\
          &   & \textbf{$\pm$0.4} &\textbf{$\pm$0.9}  &\textbf{ $\pm$0.2} & \textbf{$\pm$0.2}  & \textbf{$\pm$0.4}  & \textbf{$\pm$0.2}  & \textbf{$\pm$0.5}  & \textbf{$\pm$0.8} & $\pm$0.7& \textbf{$\pm$0.3}  & \textbf{$\pm$0.4} & $\pm$0.4 \\
		
		\midrule
		FRCN \cite{faster}  & \checkmark     & 3.26  & 6.66  & 3.04  & 3.73  & 7.79  & 3.22  & 4.59  & 9.52  & 4.07  & 5.32  & 11.20  & 4.54 \\
		
		TFA$_{\mathrm{fc}}$ \cite{wang2020frustratingly} & \checkmark     & 2.78  & 5.39  & 2.36  & 4.14  & 7.98  & 4.01  & 6.33  & 12.10  & 5.94  & 7.92 & 15.58 & 7.29 \\
		
	    TFA$_{\mathrm{cos}}$ \cite{wang2020frustratingly} & \checkmark     & 3.09  & 5.24  & 3.21  & 4.21  & 7.70   & 4.35  & 6.05  & 11.48 & 5.93  & 7.61 & 14.56 & 7.17 \\
		
		FSDetView \cite{xiao2020few} & \checkmark     & 2.20   & 6.20   & 0.90   & 3.40   & 10.00 & 1.50  & 5.20   & 14.70  & 2.10   & 8.20   & \red{21.60} & 4.70 \\
		
		MPSR \cite{wu2020multi} & \checkmark     & 3.34 & 6.11  & 3.25  & 5.41 & 9.68  & 5.52 & 5.70   & 10.54 & 5.50   & 7.20   & 13.55 & 6.89 \\
		
		A-RPN \cite{fan2020few} & \checkmark     & {4.59}  & {8.85} & {4.37} & {6.15}  & {12.05}  & {5.76}  & {8.24}   & {15.52} & {7.92} & {9.02}  & 17.29 & {8.53} \\
		
		W. Zhang \etal \cite{zhang2021hallucination} & \checkmark     & {4.40}  & {7.50} & {4.90} & {5.60}  & {9.90}  & {5.90}  & {7.20}   & {13.30} & 7.40 & -  & - & - \\
		
		FADI \cite{cao2021nips} & \checkmark     & \green{5.70} & \green{10.40}  & \green{6.00}  & \green{7.00} & \green{13.01}  & \green{7.00} & \green{8.60}   & \green{15.80} & \green{8.30}   & \green{10.10}   & 18.60 & \red{11.90} \\
		
		\textbf{AirDet (Ours)} & \checkmark     & \textbf{\red{6.10}} & \textbf{\red{11.40}} & \textbf{\red{6.04}} & \textbf{\red{8.73}} & \textbf{\red{16.24}} & \textbf{\red{8.35}} & \textbf{\red{9.95}} & \textbf{\red{19.39}} & \textbf{\red{9.09}} & \textbf{\red{10.81}} &\green{\textbf{20.75}} & \textbf{\green{10.27}} \\
	
		\bottomrule
	\end{tabular}\label{tab:coco}%
	\end{threeparttable}
\end{table}%
\myparagraph{Overall Performance}
As shown in \tref{tab:coco}, AirDet achieves significant performance gain on the baseline \cite{fan2020few}.
AirDet without fine-tuning amazingly also achieves comparable or even better results than many fine-tuned methods.
With fine-tuning, AirDet outperformed existing SOTAs \cite{fan2020few,wang2020frustratingly,xiao2020few,wu2020multi,faster,cao2021nips,zhang2021hallucination}. 
Since the results without fine-tuning may be sensitive to support images, we report the averaged performance, and the standard deviation on 3-5 randomly sampled support images, where we surprisingly find AirDet more robust to the variance of support images compared with the baseline \cite{fan2020few}.

\myparagraph{Multi-scale Objects} 
\begin{table*}[!t]
	\centering
	\setlength{\tabcolsep}{0.2mm}
	\fontsize{5.5}{6.5}\selectfont
	\caption{Performance evaluation on multi-scale objects from COCO. Highest-ranking and second-best scores are marked out with \red{red} and \green{green}, respectively. Without fine-tuning, AirDet can avoid over-fitting and shows robustness on small-scale objects. By virtue of the SCS module, AirDet can achieve higher results than those with FPN.}
	\begin{threeparttable}
		\begin{tabular}{ccc|ccc|ccc|ccc|ccc}
			\toprule
			\multicolumn{3}{c|}{Shots} & \multicolumn{3}{c|}{1} & \multicolumn{3}{c|}{2} & \multicolumn{3}{c|}{3} & \multicolumn{3}{c}{5} \\
			\midrule
			Method & FPN & Fine-tune & AP$_s$ & AP$_m$ & AP$_l$ & AP$_s$ & AP$_m$ & AP$_l$ & AP$_s$ & AP$_m$ & AP$_l$ & AP$_s$ & AP$_m$ & AP$_l$   \\
			\multirow{2}{*}{A-RPN \cite{fan2020few}}$\dag$ & \multirow{2}{*}{\xmark} & \multirow{2}{*}{\xmark}  & 2.43 & 5.00  & 6.74 & 2.67 & 5.01   & 7.18  & 3.42 & 6.15  & 8.77  & 3.54 & 6.73  & 9.97 \\
			
            &  &   & $\pm$0.4 &$\pm$1.0  & $\pm$1.1 & $\pm$0.3  & $\pm$0.3  & $\pm$0.4  & $\pm$0.2  & $\pm$0.5  & $\pm$0.8 & $\pm$0.3  & $\pm$0.03 & $\pm$0.2 \\
			\midrule
			\multirow{2}{*}{\textbf{AirDet (Ours)}}$\dag$  & \multirow{2}{*}{\xmark}     & \multirow{2}{*}{\xmark} & \red{\textbf{2.85}} & \red{\textbf{6.33}} & \red{\textbf{9.00}} & \red{\textbf{4.00}} & \red{\textbf{6.84}} & \red{\textbf{9.94}} & \red{\textbf{4.13}} & \red{\textbf{7.95}} & \red{\textbf{11.30}} & \red{\textbf{4.22}} & \red{\textbf{8.24}} & \red{\textbf{12.90}} \\
			
             &  &   & \textbf{$\pm$0.3} &\textbf{$\pm$0.7}  &\textbf{ $\pm$0.8} & \textbf{$\pm$0.3}  & \textbf{$\pm$0.1}  & \textbf{$\pm$0.3}  & \textbf{$\pm$0.1}  & \textbf{$\pm$0.5} & $\pm$0.9 & \textbf{$\pm$0.2}  & $\pm$0.04 & $\pm$0.5 \\
			
			\midrule
			FRCN \cite{faster}  & \checkmark & \checkmark     & 1.05  & 3.68  & 5.41  & 0.94  & 4.39  & 6.42  & 1.12  & 5.11  & 7.83  & 1.99  & 5.30   & 8.84 \\
			
			TFA$_{\mathrm{fc}}$ \cite{wang2020frustratingly} & \checkmark & \checkmark     & 1.06  & 2.71  & 4.38  & 1.17  & 4.02  & 7.05  & 1.97  & 5.48  & 11.09 & 2.40 & 6.86 & 12.86 \\
			
			TFA$_{\mathrm{cos}}$ \cite{wang2020frustratingly} & \checkmark & \checkmark  & 1.07  & 2.78  & 5.12  & 1.64  & 4.12  & 7.27  & 2.34  & 5.48  & 10.43 & 2.82  & 6.70   & 12.21 \\
			
			FSDetView \cite{wang2020frustratingly} & \xmark  & \checkmark  & 0.70   & 2.70   & 3.70   & 0.60   & 4.00     & 4.20   & 1.80   & 5.10   & 8.00     & \green{3.00} & 9.70   & 12.30 \\
			
			MPSR \cite{wu2020multi} & \checkmark & \checkmark & 1.23  & 3.82  & 5.58 & 1.89  & 5.69  & 8.73 & 0.86  & 4.60   & 9.96  & 1.62  & 6.78  & 11.66 \\
			
			A-RPN \cite{fan2020few} & \xmark  & \checkmark & \green{1.74} & \green{5.17} & \green{6.96}  & \green{2.20} & \green{7.55} & \green{10.49}  & \green{2.72} & \green{9.51} & \green{14.74} & 2.92  & \green{10.67} & \green{16.08} \\
			
			\textbf{AirDet (Ours)} & \xmark  & \checkmark & \red{\textbf{3.05}} & \red{\textbf{6.40}} & \red{\textbf{10.03}} & \red{\textbf{4.00}} & \red{\textbf{9.65}}     & \red{\textbf{13.91}}    & \red{\textbf{3.46}}   & \red{\textbf{11.44}}   & \red{\textbf{16.04}} & \red{\textbf{3.27}} & \red{\textbf{11.20}}    & \red{\textbf{18.64}} \\
			\bottomrule
		\end{tabular}\label{tab:coco_scale}%
	\end{threeparttable}
\end{table*}%
We next report the performance of methods \cite{wang2020frustratingly,xiao2020few,wu2020multi,fan2020few,faster} and AirDet on multi-scale objects in \tref{tab:coco_scale}.
Thanks to SCS, AirDet achieves the highest performance for multi-scale objects among all the SOTAs.
Especially for small objects, given 5-shots, AirDet can achieve a surprising \textbf{4.22} AP$_s$, nearly doubling the fine-tuned methods with multi-scale FPN features \cite{wang2020frustratingly,wu2020multi}. 

\myparagraph{Comparison of 10-Shot} 
\begin{table*}[!t]
	\centering
	\setlength{\tabcolsep}{0.1mm}
	\fontsize{5}{6.5}\selectfont
	\caption{Performance comparison with 10-shot on COCO validation dataset. \red{Red} and \green{green} fonts indicate best and second-best scores, respectively. AirDet achieves comparable results without fine-tuning and outperforms most methods with fine-tuning, which strongly demonstrates its effectiveness.}
	\begin{threeparttable}
    \begin{tabular}{ccccccccccccccc}
    \toprule[1pt]
    Method & Venue & Fine-tune & AP    & AP$_{50}$  & AP$_{75}$  & AP$_s$   & AP$_m$   & AP$_l$   & AR$_{1}$   & AR$_{10}$  & AR$_{100}$ & AR$_s$   & AR$_m$   & AR$_l$ \\
    \midrule
    LSTD \cite{chen2018lstd}  & AAAI 2018 & \checkmark  & 3.2   & 8.1   & 2.1   & 0.9   & 2.0   & 6.5   & 7.8   & 10.4  & 10.4  & 1.1   & 5.6   & 19.6 \\
    MetaDet \cite{wang2019meta} & ICCV 2019 & \checkmark & 7.1   & 14.6  & 6.1   & 1.0   & 4.1   & 12.2  & 11.9  & 15.1  & 15.5  & 1.7   & 9.7   & 30.1 \\
    FSRW \cite{kang2019few}  & ICCV 2019 & \checkmark & 5.6   & 12.3  & 4.6   & 0.9   & 3.5   & 10.5  & 10.1  & 14.3  & 14.4  & 1.5   & 8.4   & 28.2 \\
    Meta RCNN \cite{yan2019meta}& ICCV 2019 & \checkmark  & 8.7   & 19.1  & 6.6   & 2.3   & 7.7   & 14.0  & 12.6  & 17.8  & 17.9  & 7.8   & 15.6  & 27.2 \\
    TFA$_{\mathrm{fc}}$ \cite{wang2020frustratingly} & ICML 2020 & \checkmark & 9.1   & 17.3  & 8.5   & -     & -     & -     & -     & -     & -     & -     & -     & - \\
    TFA$_{\mathrm{cos}}$ \cite{wang2020frustratingly} & ICML 2020 & \checkmark & 9.1   & 17.1  & 8.8   & -     & -     & -     & -     & -     & -     & -     & -     & - \\
    FSDetView \cite{xiao2020few}& ECCV 2020 & \checkmark  & 12.5  & \red{27.3}  & 9.8   & 2.5   & 13.8  & 19.9  & 20.0  & 25.5  & 25.7  & 7.5   & 27.6  & 38.9 \\
    MPSR \cite{wu2020multi} & ECCV 2020 & \checkmark & 9.8   & 17.9  & 9.7   & 3.3   & 9.2   & 16.1  & 15.7  & 21.2  & 21.2  & 4.6   & 19.6  & 34.3 \\
    A-RPN \cite{fan2020few} & CVPR 2020 & \checkmark & 11.1  & 20.4  & 10.6  & -     & -     & -     & -     & -     & -     & -     & -     & - \\
    SRR-FSD \cite{zhu2021semantic}& CVPR 2021 & \checkmark & 11.3  & 23.0  & 9.8   & -     & -     & -     & -     & -     & -     & -     & -     & - \\
    FSCE \cite{sun2021fsce} & CVPR 2021 & \checkmark  & 11.9  & -     & 10.5  & -     & -     & -     & -     & -     & -     & -     & -     & - \\
    DCNet \cite{Hu2021CVPR} & CVPR 2021 & \checkmark & \green{12.8}  & 23.4  & 11.2  & 4.3   & \green{13.8}  & \green{21.0}  & 18.1  & 26.7  & 25.6  & 7.9   & 24.5  & 36.7 \\
    Y. Li \etal \cite{li2021few} & CVPR 2021 & \checkmark  & 11.3  & 20.3   & - & -  & -   & - & -     & -     & -     & -     & -     & - \\
    FADI \cite{cao2021nips} & NIPS 2021 & \checkmark  & 12.2 & 22.7  & \green{11.9}  & -     & -     & -     & -     & -     & -     & -     & -     & - \\
    QA-FewDet \cite{han2021query} & ICCV 2021 & \checkmark  & 11.6  & 23.9  & 9.8  & -     & -     & -     & -     & -     & -     & -     & -     & - \\
    FSOD$^{up}$ \cite{wu2021universal} & ICCV 2021 & \checkmark  & 11.0  & -     & 10.7  & \red{4.5}   & 11.2  & 17.3  & -     & -     & -     & -     & -     & - \\
    \midrule
    \textbf{AirDet} & \textbf{Ours} & \xmark   & 8.7    & 15.3  & 8.8  & 4.3     & 9.7   & 14.8   & \green{\textbf{19.1}}   & \red{\textbf{33.8}}   & \red{\textbf{34.8}}   & \red{\textbf{13.0}}    & \red{\textbf{37.4}}   & \green{\textbf{52.9}} \\
    \textbf{AirDet} & \textbf{Ours} & \checkmark & \red{\textbf{13.0}}     & \green{\textbf{23.9}}   & \red{\textbf{12.4}}   & \red{\textbf{4.5}}   & \red{\textbf{15.2}}   & \red{\textbf{22.8}}   & \red{\textbf{20.5}}     & \green{\textbf{33.7}} & \green{\textbf{34.4}}  & \green{\textbf{9.6}}  & \green{\textbf{36.4}}   & \red{\textbf{55.0}} \\
    \bottomrule[1pt]
	\end{tabular}\label{tab:10shot}%
	\end{threeparttable}
\end{table*}%
For a more thorough comparison, we present the 10-shot evaluation on the COCO validation dataset in \tref{tab:10shot}. Without fine-tuning, AirDet can surprisingly achieve comparable performance against recent work \cite{wu2020multi,wang2020frustratingly,yan2019meta}, while all of them require a careful fine-tuning stage. Moreover, our fine-tuned model outperforms most prior methods \cite{li2021few,cao2021nips,han2021query,wu2021universal,Hu2021CVPR,sun2021fsce,zhu2021semantic,fan2020few,wu2020multi,xiao2020few,wang2020frustratingly,yan2019meta,kang2019few,wang2019meta,chen2018lstd} in most metrics, especially average recall rate (AR). Besides, the performance superiority on small objects (AP$_s$ and AR$_s$) further demonstrates the effectiveness of AirDet on multi-scale, especially the small-scale objects.

\myparagraph{Efficiency Comparison}
We report the fine-tuning and inference time of AirDet, and the SOTA methods \cite{fan2020few,wang2020frustratingly,xiao2020few,wu2020multi} in a setting of 3-shot one class in \tref{tab:time}, in which the official code and implementation with ResNet101 as the backbone are adopted.
Without fine-tuning, AirDet can make direct inferences on novel objects with a comparable speed, while the others methods \cite{fan2020few,xiao2020few,wu2020multi,wang2020frustratingly} require a fine-tuning time of about 3-30 minutes, which cannot meet the requirements of online exploration.
Note that the fine-tuning time is measured on TITAN X GPU, while such computational power is often unavailable on robots.

\Remark Many methods \cite{li2021few,cao2021nips,han2021query,wu2021universal,Hu2021CVPR,sun2021fsce,zhu2021semantic,fan2020few,wu2020multi,xiao2020few,wang2020frustratingly,yan2019meta,kang2019few,wang2019meta,chen2018lstd} also require an offline process to fine-tune hyper-parameters for different shots.
While such \textit{off-line} tuning is infeasible for robotic \textit{online} exploration. 
Instead, AirDet can adopt \textbf{the same} base-trained model without fine-tuning for implementation.

\begin{table}[!t]
	\centering
	\setlength{\tabcolsep}{0.1mm}
	 \caption{Efficiency comparison with official source code. We adopt the pre-trained models provided by \cite{wang2020frustratingly}, so their fine-tuning time is unavailable.}
	\fontsize{5}{7.5}\selectfont
    \begin{tabular}{cccccccc}
    \toprule
    Method & \textbf{AirDet} & \multicolumn{1}{l}{A-RPN} \cite{fan2020few} & \multicolumn{1}{l}{FSDet} \cite{xiao2020few}& \multicolumn{1}{l}{MPSR} \cite{wu2020multi} & \multicolumn{1}{l}{TFA$_{\mathrm{fc}}$} \cite{wang2020frustratingly} & \multicolumn{1}{l}{TFA$_{\mathrm{cos}}$} \cite{wang2020frustratingly} & \multicolumn{1}{l}{FRCN$_{\mathrm{ft}}$} \cite{wang2020frustratingly} \\
    \midrule
    Fine-tuning (min) & \textbf{0} & 21    & 11    & 3     & - & - & - \\
    Inference (s/img) & \textbf{0.081} & 0.076 & 0.202 & 0.109 & 0.085 & 0.094 & 0.091 \\
    \bottomrule
    \end{tabular}%
  \label{tab:time}%
\end{table}%

\begin{table}[!t]
	\centering
	\setlength{\tabcolsep}{0.2mm}
	\fontsize{5.5}{6.5}\selectfont
	\caption{Cross-domain performance on VOC-2012 validation dataset. \red{Red} and \green{green} fonts denote the first and second place, respectively. AirDet has been demonstrated strong generalization capability, maintaining obvious superiority against others.}
	\begin{threeparttable}
	\begin{tabular}{cc|ccc|ccc|ccc|ccc}
		\toprule
		\multicolumn{2}{c|}{Shots} & \multicolumn{3}{c|}{1} & \multicolumn{3}{c|}{2} & \multicolumn{3}{c|}{3} & \multicolumn{3}{c}{5} \\
		\midrule
		Method  & Fine-tune & AP & AP$_{50}$ & AP$_{75}$ & AP & AP$_{50}$ & AP$_{75}$ & AP & AP$_{50}$ & AP$_{75}$ & AP & AP$_{50}$ & AP$_{75}$  \\
		\multirow{2}{*}{A-RPN \cite{fan2020few}}$\dag$ & \multirow{2}{*}{\xmark}    & 10.45  & 18.10 & 10.32 & 13.10  & 22.60  & 13.17  & 14.05  & 24.08 & 14.24 & 14.87  & 25.03 & 15.26 \\
		
          &   & $\pm$0.1 &$\pm$0.1  & $\pm$0.1 & $\pm$0.2  & $\pm$0.4  & $\pm$0.2  & $\pm$0.2  & $\pm$0.2  & $\pm$0.2 & $\pm$0.08  & $\pm$0.07 & $\pm$0.1 \\
        \midrule
		\multirow{2}{*}{\textbf{AirDet (Ours)}}$\dag$  & \multirow{2}{*}{\xmark}     & \red{\textbf{11.92}} & \red{\textbf{21.33}} & \red{\textbf{11.56}} & \red{\textbf{15.80}} & \red{\textbf{26.80}} & \red{\textbf{16.08}} & \red{\textbf{16.89}} & \red{\textbf{28.61}} & \red{\textbf{17.36}} & \red{\textbf{17.83}} & \red{\textbf{29.78}} & \red{\textbf{
	18.38}} \\
          &   & \textbf{$\pm$0.06} &\textbf{$\pm$0.08}  &\textbf{ $\pm$0.08} & \textbf{$\pm$0.08}  & \textbf{$\pm$0.3}  & \textbf{$\pm$0.05}  & \textbf{$\pm$0.1}  & \textbf{$\pm$0.1} & \textbf{$\pm$0.1} & \textbf{$\pm$0.03}  & \textbf{$\pm$0.03} & \textbf{$\pm$0.1} \\
		
		\midrule
		FRCN \cite{faster} & \checkmark   & \multicolumn{1}{c}{4.49} & \multicolumn{1}{c}{9.44} & \multicolumn{1}{c|}{3.85} & \multicolumn{1}{c}{5.20} & \multicolumn{1}{c}{11.92} & \multicolumn{1}{c|}{3.84} & \multicolumn{1}{c}{6.50} & \multicolumn{1}{c}{14.39} & \multicolumn{1}{c|}{5.11} & \multicolumn{1}{c}{6.55} & \multicolumn{1}{c}{14.48} & \multicolumn{1}{c}{5.09} \\
		
		TFA$_{\mathrm{cos}}$ \cite{wang2020frustratingly} & \checkmark   & \multicolumn{1}{c}{4.66} & \multicolumn{1}{c}{7.97} & \multicolumn{1}{c|}{5.14} & \multicolumn{1}{c}{6.59} & \multicolumn{1}{c}{11.91} & \multicolumn{1}{c|}{6.49} & \multicolumn{1}{c}{8.78} & \multicolumn{1}{c}{17.09} & \multicolumn{1}{c|}{8.15} & \multicolumn{1}{c}{10.46} & \multicolumn{1}{c}{20.93} & \multicolumn{1}{c}{9.53} \\
		
		TFA$_{\mathrm{fc}}$ \cite{wang2020frustratingly} & \checkmark  & \multicolumn{1}{c}{4.40} & \multicolumn{1}{c}{8.60} & \multicolumn{1}{c|}{4.21} & \multicolumn{1}{c}{7.02} & \multicolumn{1}{c}{13.80} & \multicolumn{1}{c|}{6.21} & \multicolumn{1}{c}{9.24} & \multicolumn{1}{c}{18.48} & \multicolumn{1}{c|}{8.03} & \multicolumn{1}{c}{11.11} & \multicolumn{1}{c}{22.83} & \multicolumn{1}{c}{9.78} \\
		
		FSDetView \cite{xiao2020few} & \checkmark  & 4.80     & 14.10    &  1.40   & 3.70   & 11.60  & 0.60   & 6.60  & 22.00  & 1.20  & 10.80  & 26.50   & 5.50 \\
		
		MPSR \cite{wu2020multi} & \checkmark & \multicolumn{1}{c}{6.01} & \multicolumn{1}{c}{11.23} & \multicolumn{1}{c|}{5.74} & \multicolumn{1}{c}{8.20} & \multicolumn{1}{c}{15.08} & \multicolumn{1}{c|}{8.22} & \multicolumn{1}{c}{10.08} & \multicolumn{1}{c}{18.29} & \multicolumn{1}{c|}{9.99} & \multicolumn{1}{c}{11.49} & \multicolumn{1}{c}{21.33} & \multicolumn{1}{c}{11.06} \\
		
		A-RPN \cite{fan2020few} & \checkmark & \green{9.49} & \green{17.41} & \green{9.42} & \green{12.71} & \green{23.66} & \green{12.44} & \green{14.89} & \green{26.30} & \green{14.76} & \green{15.09} & \green{28.08} & \green{14.17} \\
		\textbf{AirDet (Ours)} & \checkmark  & \textbf{\red{13.33}}     & \textbf{\red{24.64}}     & \textbf{\red{12.68}}   & \textbf{\red{17.51}}    & \textbf{\red{30.35}}   & \textbf{\red{17.61}}   & \textbf{\red{17.68}}   & \textbf{\red{32.05}}  & \textbf{\red{17.34}}   & \textbf{\red{18.27}}  & \textbf{\red{33.02}} & \textbf{\red{17.69}} \\
		\bottomrule[1.2pt]
	\end{tabular}\label{tab:voc}%
	\end{threeparttable}
\end{table}%

\subsection{Cross-domain Evaluation}\label{sec:cross}

Robots are often deployed to novel environments that have never been seen during training, thus cross-domain test is crucial for robotic applications.
In this section, we adopt the same model trained on COCO, while test on PASCAL VOC \cite{everingham2010pascal} and LVIS \cite{gupta2019lvis} to evaluate the model generalization capability.

\myparagraph{PASCAL VOC}
We report the overall performance on PASCAL VOC-2012 \cite{everingham2010pascal} for all methods in \tref{tab:voc}.
In the cross-domain setting, even without fine-tuning, AirDet achieves better performance than methods \cite{wu2020multi,fan2020few,faster,xiao2020few,wang2020frustratingly} that perform relatively well in in-domain test. 
This means AirDet has a much stronger generalization capability than most fine-tuned prior methods. 

\myparagraph{LVIS}
We randomly sample LVIS \cite{gupta2019lvis} to form 4 splits of classes, each of which contains 16 different classes. 
To provide valid evaluation, the classes that have 20 to 200 images are taken for the test.
More details can be found in \appref{sec:lvis}.
The averaged performance with 5-shot without fine-tuning is presented in \tref{tab:lvis-cross}, where AirDet outperforms the baseline \cite{fan2020few} in every split under all metrics. Since the novel categories in the 4 LVIS splits are more (64 classes in total) and rarer (many of them are uncommon) than the VOC 20 classes, the superiority of AirDet in \tref{tab:lvis-cross} highly demonstrate its robustness under class variance.


\begin{table}[!t]
	\setlength{\tabcolsep}{.6mm}
	\centering
    \fontsize{5.5}{6.5}\selectfont
  \caption{Cross-domain performance of A-RPN \cite{fan2020few} and AirDet on LVIS dataset. We report the results for 5-shot without fine-tuning on 4 random splits.}
    \begin{tabular}{c|cccc|cccc|cccc|cccc}
    \toprule
    \multicolumn{1}{c|}{Split} & \multicolumn{4}{c|}{1}        & \multicolumn{4}{c|}{2}        & \multicolumn{4}{c|}{3}        & \multicolumn{4}{c}{4} \\
    \midrule
    Metrict & AP & AP$_{50}$ & AP$_{75}$    & AR$_{10}$    & AP & AP$_{50}$ & AP$_{75}$    & AR$_{10}$    & AP & AP$_{50}$ & AP$_{75}$   & AR$_{10}$    & AP & AP$_{50}$ & AP$_{75}$  & AR$_{10}$ \\
    \textbf{AirDet} & \textbf{6.71} & \textbf{12.31} & \textbf{6.51} & \textbf{27.57} & \textbf{9.35} & \textbf{14.23} & \textbf{9.98} & \textbf{25.42} & \textbf{9.09} & \textbf{15.64} & \textbf{8.82} & \textbf{34.64} & \textbf{11.07} & \textbf{16.90} & \textbf{12.30} & \textbf{25.76} \\
    A-RPN & 5.49  & 10.04 & 5.27  & 26.59 & 8.85  & 13.41 & 9.46  & 24.45 & 7.49  & 12.34 & 8.13  & 33.85 & 10.80  & 15.46 & 12.24 & 25.05 \\
    \bottomrule
    \end{tabular}%
  \label{tab:lvis-cross}%
\end{table}%

\subsection{Ablation Study and Deep Visualization}\label{sec:abla}
In this section, we address the effectiveness of the proposed three modules via quantitative results and qualitative visualization using Grad-Cam \cite{gradcam}.

\myparagraph{Quantitative Evaluation}
We report the overall performance on 3-shot and 5-shot for the baseline \cite{fan2020few} and AirDet by enabling the three modules, respectively.
It can be seen in \tref{tab:ABLA} that AirDet outperforms the baseline in all cases. With the modules enabled one by one, the results get gradually higher, which strongly demonstrates the necessity and effectiveness of SCS, GLR, and PRE.

\begin{table}[!t]
	\centering
	\setlength{\tabcolsep}{0.2mm}
	\fontsize{5.5}{6.5}\selectfont
	\caption{Ablation study of the three modules, \ie, PRE, GLR, and SCS in AirDet. With each module enabled, the performance is improved step by step on our baseline. With the full modules, AirDet can amazingly achieve up to \textbf{35\%} higher results.} 
    \begin{tabular}{ccc|cccccc|cccccc}
    \toprule
    \multicolumn{3}{c|}{Module} & \multicolumn{5}{c}{3}                 &       & \multicolumn{6}{c}{5} \\
    \midrule
    PRE & GLR   & SCS   & \multicolumn{1}{c}{AP} & $\Delta\%$ & \multicolumn{1}{c}{AP$_{50}$} & $\Delta\%$ & \multicolumn{1}{c}{AP$_{75}$} & $\Delta\%$ & \multicolumn{1}{c}{AP} & $\Delta\%$ & \multicolumn{1}{c}{AP$_{50}$} & $\Delta\%$ & \multicolumn{1}{c}{AP$_{75}$} & $\Delta\%$ \\
    \multicolumn{3}{c|}{Baseline \cite{fan2020few} } & 4.80  & 0.00  & 9.24  & 0.00  & 4.49  & 0.00  & 5.73  & 0.00  & 10.68 & 0.00  & 5.53  & 0.00 \\
    \checkmark     & \multicolumn{1}{c}{} & \multicolumn{1}{c|}{} & 5.15  & +7.29 & 10.11 & +9.41 & 4.71  & +4.90 & 5.94  & +3.66  & 11.54 & +8.05  & 5.34  & -3.43 \\
    \checkmark     & \checkmark     & \multicolumn{1}{c|}{} & 5.59  & +16.46 & 10.61 & +14.83 & 5.12  & +14.03 & 6.44  & +12.39 & 12.08 & +13.11 & 6.06  & +9.58 \\
    \midrule
    \checkmark     & \checkmark     & \checkmark     & \textbf{6.50} & \textbf{+35.41} & \textbf{12.30} & \textbf{+33.12} & \textbf{6.11} & \textbf{+36.08} & \textbf{7.27} & \textbf{+26.78} & \textbf{13.63} & \textbf{+27.62} & \textbf{6.71} & \textbf{+21.34} \\
    \bottomrule
    \end{tabular}%
  \label{tab:ABLA}%
\end{table}%

\begin{figure}[!t]
	\centering
	\includegraphics[width=1\columnwidth]{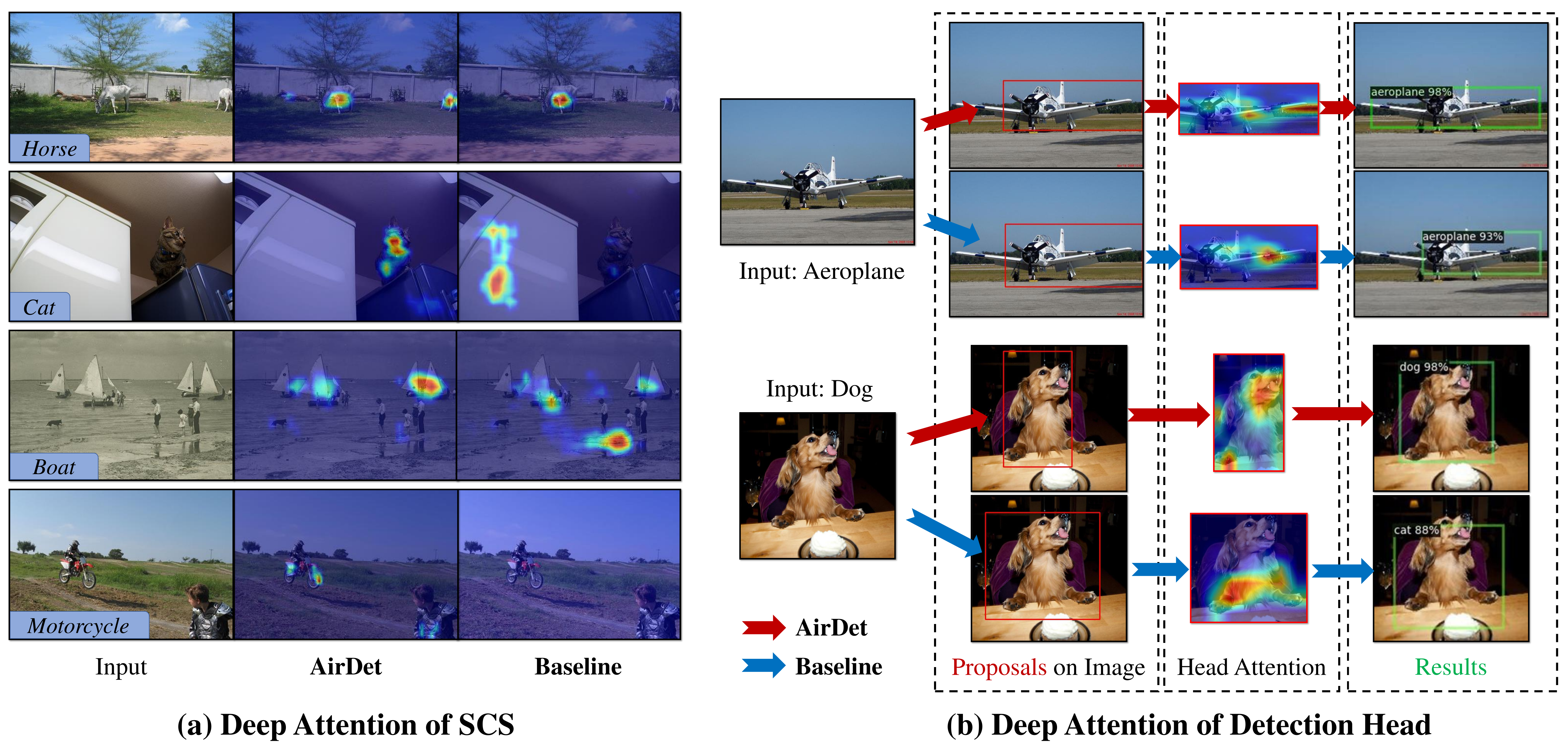}
	\caption{Deep visualization comparison between AirDet and baseline \cite{fan2020few}. In (a), By virtue of SCS, AirDet is capable of finding given support objects effectively. In (b), with similar proposals (\textcolor[rgb]{1,0,0}{red} boxes), AirDet can focus on the entire object (aeroplane) and notice the most representative parts (dog), resulting in more precise regression box and correct classification results. More examples are presented in \appref{sec:more_deep}.}
	\label{fig:deep_rpn}
\end{figure}

\myparagraph{How effective is SCS?} 
Given 2-shot per class, we first take the highest ranking proposal from RPN \cite{faster} to backpropagate the objectiveness score and resize the gradient map to the original image.
\fref{fig:deep_rpn} (a) exhibits the heat map from both AirDet and the baseline. We observe that AirDet generally concentrates on objects more precisely than the baseline.
Moreover, AirDet can focus better on objects belonging to the support class and is not distracted by other objects (2nd and 3rd row).
This means that AirDet can generate novel object proposals more effectively.

\myparagraph{How effective is GLR and detection head?}
In \fref{fig:deep_rpn} (b), we observe that with similar proposal boxes, AirDet head can better focus on the entire object, \eg, aeroplane is detected with a precise regression box, \eg, the dog is correctly classified with high score. This again demonstrates the effectiveness of our GLR and detection head.

\subsection{Real-World Test}\label{sec:real}
\begin{table*}[!t]
	\centering
	\setlength{\tabcolsep}{0.6mm}
	\fontsize{5.5}{6.5}\selectfont
	\caption{3-shot real-world exploration test of AirDet and baseline \cite{fan2020few}. AirDet can be directly applied without fine-tuning and performs considerably more robust than the baseline by virtue of the newly proposed SCS, GLR, and PRE modules.} 
    \begin{tabular}{ccc|cc|cc|cc|cc|cc}
    \toprule
    \multicolumn{13}{c}{Real-world Exploration Test} \\
    \midrule
    \multicolumn{1}{l}{Test/\#Frames} & \multicolumn{2}{c|}{1/\#248} & \multicolumn{2}{c|}{2/\#146} & \multicolumn{2}{c|}{3/\#127} & \multicolumn{2}{c|}{4/\#41} & \multicolumn{2}{c|}{5/\#248} & \multicolumn{2}{c}{6/\#46} \\
    \midrule
    Metric & AP    & AP$_{50}$  & AP    & AP$_{50}$ & AP    & AP$_{50}$  & AP    & AP$_{50}$  & AP    & AP$_{50}$  & AP    & AP$_{50}$ \\
    \textbf{AirDet (Ours)} & \textbf{17.10} & \textbf{54.10} & \textbf{17.90} & \textbf{47.40} & \textbf{24.00} & \textbf{57.50} & \textbf{26.94} & \textbf{48.20} & \textbf{11.28} & \textbf{38.17} & \textbf{20.40} & \textbf{70.63} \\
    A-RPN \cite{fan2020few} & 13.56 & 40.40 & 14.30 & 38.80 & 20.20 & 47.20 & 22.41 & 40.14 & 6.75  & 24.10 & 14.70 & 59.38 \\
    \midrule
    Test/\#Frames & \multicolumn{2}{c|}{7/\#212} & \multicolumn{2}{c|}{8/\#259} & \multicolumn{2}{c|}{9/\#683} & \multicolumn{2}{c|}{10/\#827} & \multicolumn{2}{c|}{11/\#732} & \multicolumn{2}{c}{12/\#50} \\
    \midrule
    Metric & AP    & AP$_{50}$  & AP    & AP$_{50}$ & AP    & AP$_{50}$  & AP    & AP$_{50}$  & AP    & AP$_{50}$  & AP    & AP$_{50}$ \\
    \textbf{AirDet (Ours)} & \textbf{5.90}  & \textbf{16.00} & \textbf{15.26} & \textbf{43.31} & \textbf{7.63}  &\textbf{27.88} & \textbf{13.55} & \textbf{23.92} & \textbf{15.74} & \textbf{34.43} & \textbf{21.45} & \textbf{45.83} \\
    A-RPN \cite{fan2020few} & 2.39  & 7.60  & 11.27 & 25.24 & 6.16  & 23.40 & 8.10  & 14.85 & 11.54 & 27.28 & 18.20 & 33.98 \\
    \bottomrule
    \end{tabular}%
  \label{tab:subt}%
\end{table*}%

Real-world tests are conducted for AirDet and our baseline \cite{fan2020few} with 12 sequences that were collected from the DARPA Subterranean (SubT) challenge \cite{subtchallenge}.
Due to the requirements of \textit{online} response during the mission, the models can only be evaluated \textbf{without fine-tuning}, which makes existing methods \cite{li2021few,cao2021nips,han2021query,wu2021universal,Hu2021CVPR,sun2021fsce,zhu2021semantic,fan2020few,wu2020multi,xiao2020few,wang2020frustratingly,yan2019meta,kang2019few,wang2019meta,chen2018lstd} impractical.
The environments of SubT challenge also poses extra difficulties, \eg, a lack of lighting, thick smoke, dripping water, and cluttered or irregularly shaped environments, \etc~
To test the generalization capabilities, we adopt the same models of AirDet and the baseline as those evaluated in \sref{sec:indomain} and \sref{sec:cross}. The performance of 3-shot for each class is exhibited in \tref{tab:subt}, where AirDet is proved better.
The robot is equipped with an NVIDIA Jetson AGX Xavier, where our method runs at 1-2 FPS without TensorRT acceleration or other optimizations.

\begin{table}[t]
  \centering
  \setlength{\tabcolsep}{2mm}
  \fontsize{6}{6}\selectfont
  \caption{Per class results of the real-world tests. We report the instance number of each novel class along with the 3-shot AP results from AirDet and A-RPN \cite{fan2020few}. Compared with the baseline, AirDet achieves higher results for all classes.}
    \begin{tabular}{cccccccc}
    \toprule[1.2pt]
    Class & Backpack & Helmet & Rope  & Drill & Vent  & Extinguisher & Survivor \\
    \midrule
    Instances & 626   & 674   & 723   & 587   & 498   & 1386  & 205 \\
    AirDet & \textbf{32.3}  & \textbf{9.7}   & \textbf{13.9}  &\textbf{ 10.8}  & \textbf{16.2}  & \textbf{10.5}  & \textbf{10.7} \\
    Baseline \cite{fan2020few} & 26.6  & 9.7   & 6     & 9     & 14.4  & 5.6   & 9.1 \\
    \bottomrule[1.2pt]
    \end{tabular}%
  \label{tab:subt_cls}%
\end{table}%

\begin{figure}[t]
	\centering
	\includegraphics[width=1\columnwidth]{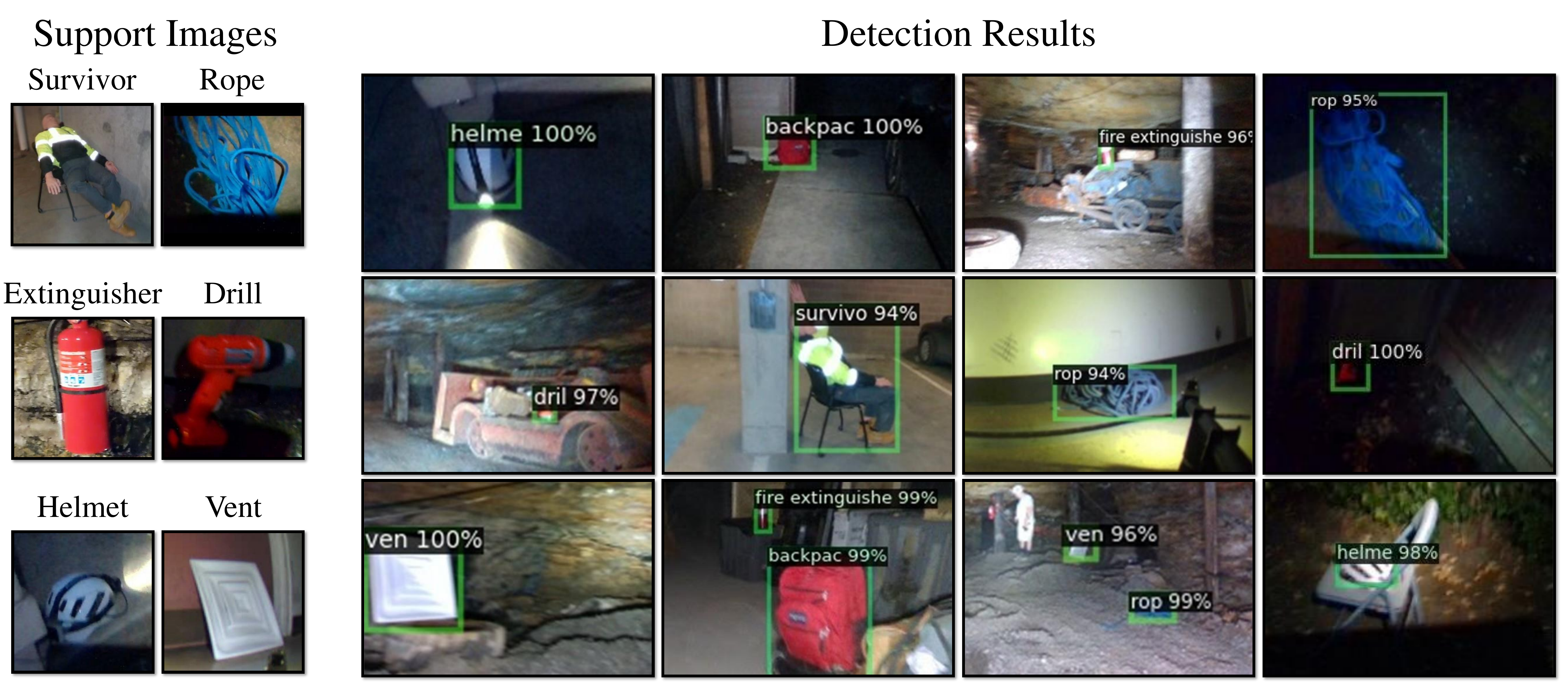}
	\caption{The provided support images and examples of detection results in the real-world tests. AirDet is robust to distinct object scales and different illumination conditions.}
	\label{fig:subt}
\end{figure}

In \tref{tab:subt_cls}, we present the number of instances and the performance on each novel class. To our excitement, AirDet shows smaller variance and higher precision cross different classes.
We also present the support images and representative detected objects in \fref{fig:subt}.
Note that AirDet can detect the novel objects accurately in the query images even if they have distinct scales and different illumination conditions with the supports. We regard this capability to the carefully designed SCS in AirDet.
More visualization are presented in \appref{sec:quali}.
The robustness and strong generalization capability of AirDet in the real-world tests demonstrated its promising prospect and feasibility for autonomous exploration.

\section{Limitation and Future Work}
Despite the promising prospect and outstanding performance, AirDet still has several limitations. 
(1) Since abundant base classes are needed to generalize, AirDet needs a relatively large base dataset to train before inference on novel classes. 
(2) Second, AirDet relies on the quality of support images to work well without fine-tuning. This is because the provided few support images are the only information for the unseen classes. 
(3) We observe that the failure cases of AirDet are mainly due to false classification, resulting in a high result variance among different classes in COCO and VOC. 
(4) Since SCS and the detection head run in loops for multiple novel classes, the efficiency of AirDet will suffer from a large number of novel classes. 
We provide quantitative results for limitation (1), (2), and (3) in \appref{sec:de_limi}.

\section{Conclusion}
This paper presents a brand new few-shot detector, AirDet, which consists of 3 newly proposed \textit{class-agnostic relation}-based modules and is free of fine-tuning.
Specifically, with proposed spatial relation and channel relation, we construct support guided cross-scale feature fusion for region proposals, global-local relation network for shots aggregation, and prototype relation embedding for precise localization. With the strong capability to extract \textit{class-agnostic relation}, AirDet can work comparably or even better than those exhaustively fine-tuned methods in both in-domain and cross-domain evaluation.
AirDet is also tested on real-world data with a robotic platform, where its feasibility for autonomous exploration is demonstrated.
\\
\par\noindent
\myparagraph{Acknowledgement}
This work was sponsored by ONR grant \#N0014-19-1-2266 and ARL DCIST CRA award W911NF-17-2-0181. The work was done when Bowen Li and Pranay Reddy were interns at The Robotics Institute, Carnegie Mellon University. The authors would like to thank all members of the Team Explorer for providing data collected from the DARPA Subterranean Challenge.

%
\bibliographystyle{splncs04}
\bibliography{egbib}

\title{Supplementary Material} 

\titlerunning{AirDet}
%
\author{Bowen Li\inst{1,2} \and
	Chen Wang\inst{1} \and
	Pranay Reddy\inst{1,3} \and\\
	Seungchan Kim\inst{1} \and
	Sebastian Scherer\inst{1}}
\authorrunning{B. Li, C. Wang, et al.}
%
\institute{Robotics Institute, Carnegie Mellon University, USA \\
	\email{chenwang@dr.com, \{bowenli2,seungch2,basti\}@andrew.cmu.edu}\\ \and
	School of Mechanical Engineering, Tongji University, China\\ \and
	Electronics and Communication Engineering, IIITDM Jabalpur, India\\
	\email{2018033@iiitdmj.ac.in}}
\maketitle
\appendix
\section*{Overview}

To ensure reproducibility, we present the detailed configuration in \appref{sec:config} and show backbone comparison in \appref{sec:backbone} for thoroughness. More qualitative results from general detection datasets, VOC-2012 validation dataset, and COCO validation dataset, as well as the representative scenes from the DARPA SubT challenge are presented in \appref{sec:quali}. We also displayed more deep visualization in \appref{sec:more_deep} to further validate the effectiveness of SCS and detection head of AirDet. Details about the LVIS dataset splits are in \appref{sec:lvis}. The limitations of AirDet are also more exhaustively studied in \appref{sec:de_limi}.

\section{Detailed Configuration}\label{sec:config}

\myparagraph{Training:} We follow our baseline \cite{fan2020few}, where contrastive training pipeline is adopted. We first reconstruct the COCO-2017 training dataset, ensuring that only one class object is annotated for each query image. During training, for one certain query image with objects belonging to class $c_1$, we provide 20 support images, including ten belonging to class $c_1$ and ten from another random class $c_2$, termed as 2-way 10-shot contrastive training. Following our baseline \cite{fan2020few}, the support images are cropped, resized, and zero-padded to $320\times320$ pixels.

\myparagraph{Inference:} Considering $N$-way $K$-shot inference, we provide all $K$-shot support data from $N$ novel classes for one query image. For each novel class, $100$ proposals are generated from the support-guided cross-scale fusion (SCS) module, and they are ranked according to the detection confidence. We finally take the top $100$ proposals in all the $N\times100$ candidates for calculating the final performance.

\myparagraph{Parameters:} The input of SCS are feature maps from ResNet2, ResNet3, and ResNet4 block. We use a global averaged support feature (weights of $\mathrm{MLP}$ and $\mathrm{Conv}$ in \eqref{eqn:inner}) are all $1$) as the $1\times 1$ convolutional kernel for multi-scale query feature. ROI Align strategy \cite{mask} is employed for pooling. The default learning rate for both 2 models with ResNet50 and ResNet101 \cite{He2016res} backbone is $0.004$. The model employing ResNet50 is trained for a total of $120,000$ iterations with ResNet1, 2, 3 blocks frozen, while the ResNet101 backbone model is trained for $80,000$ iterations with only ResNet1 block frozen. We observe that for a deeper backbone, freeing ResNet2 and 3 blocks will help the SCS module generate effective proposals better. For both the two models, the detection head takes a learning rate of $0.008$. We have maintained all other parameters the same as our baseline \cite{fan2020few}. Please refer to the attached code for more details.

\section{Backbone Comparison}\label{sec:backbone}
\begin{table}[!t]
	\centering
	\setlength{\tabcolsep}{0.6mm}
	\fontsize{6}{8.5}\selectfont
	\caption{Performance comparison of AirDet with different backbones on COCO validation dataset. The model with ResNet101 backbone performs generally better.}
    \begin{tabular}{c|rrrr|cccc|cccc|cccc}
    \toprule
    Shots & \multicolumn{4}{c|}{1}        & \multicolumn{4}{c|}{2}        & \multicolumn{4}{c|}{3}        & \multicolumn{4}{c}{5} \\
\cmidrule{2-17}    Metric & AP    & AP$_{50}$    & AP$_{75}$    & AP$_{s}$ & AP    & AP$_{50}$    & AP$_{75}$    & AP$_{s}$    & AP    & AP$_{50}$    & AP$_{75}$    & AP$_{s}$   & AP & AP$_{50}$    & AP$_{75}$    & AP$_{s}$ \\
    ResNet101 & 6.00 & 10.78 & 5.92 & 2.77  & 6.63  & 12.12 & 6.37  & 3.33  & 6.94  & 12.96 & 6.49  & 4.05  & 7.63  & 13.86 & 7.34  & 4.32 \\
    ResNet50 & 4.64  & 9.60   & 3.97  & 1.82 & 5.59  & 10.81 & 5.16  & 2.73  & 6.38  & 12.29 & 5.83  & 2.83  & 7.43  & 13.78 & 7.17  & 3.30 \\
    \bottomrule
    \end{tabular}%
  \label{tab:backbone}%
\end{table}%

\begin{figure*}[!t]
	\centering
	\includegraphics[width=1\textwidth]{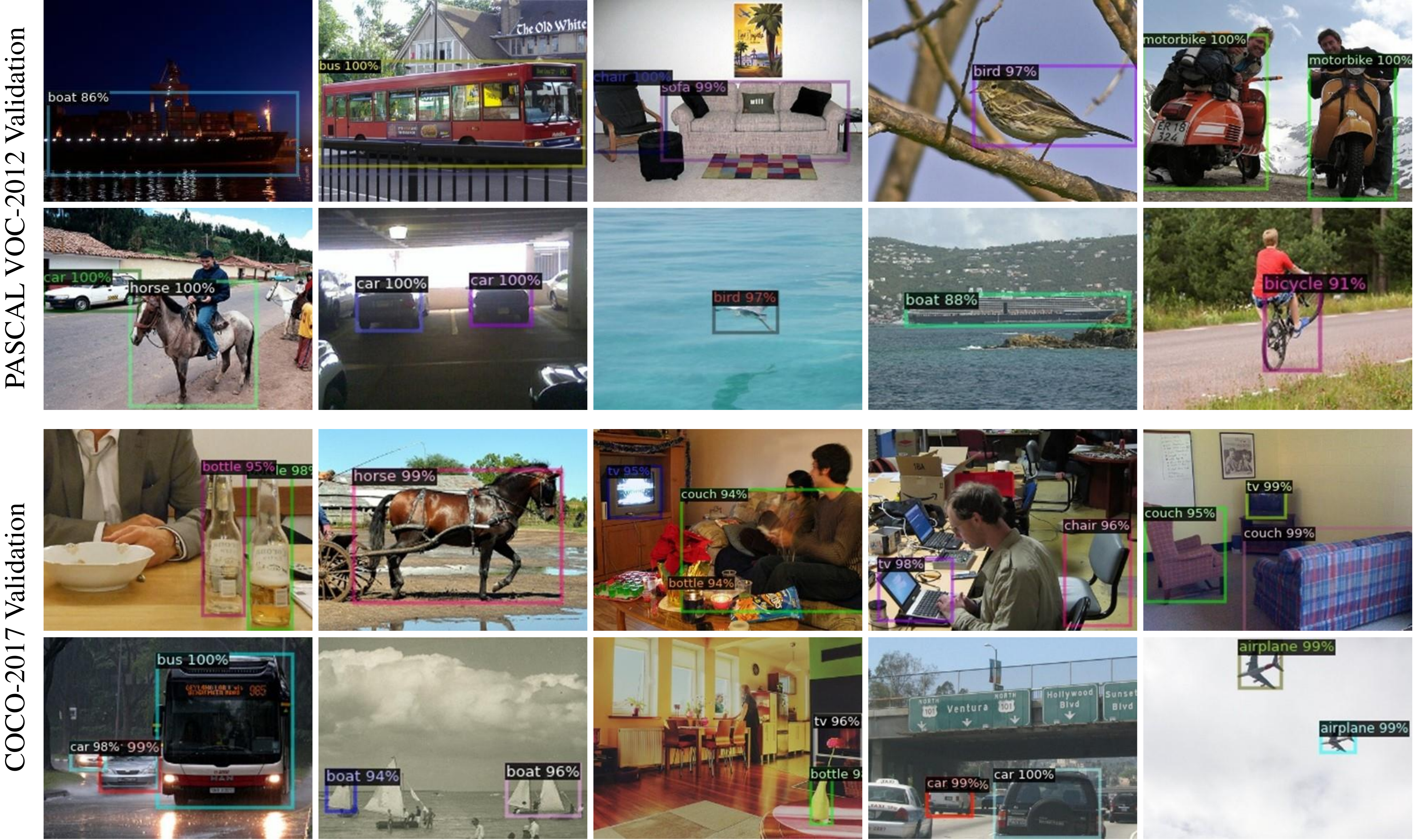}
	\caption{Representative examples of 3-shot detection of AirDet on VOC-2012 validation and COCO validation dataset. Without fine-tuning, AirDet can robustly detect the novel unseen objects such as boat, bus, sofa, with merely three support images.}
	\label{fig:voco}
\end{figure*}

The performance comparison of AirDet with different backbones \cite{He2016res} are shown in \tref{tab:backbone}. We report the average results of AirDet with ResNet101 and ResNet50 backbone on COCO validation dataset, using the same support examples. We find the model with ResNet101 generally perform better, while the switch to ResNet50 also doesn't result in too severe performance drop, which demonstrates the universal property of AirDet architecture.

\section{More Qualitative Results}\label{sec:quali}

\begin{figure*}[!t]
	\centering
	\includegraphics[width=1\textwidth]{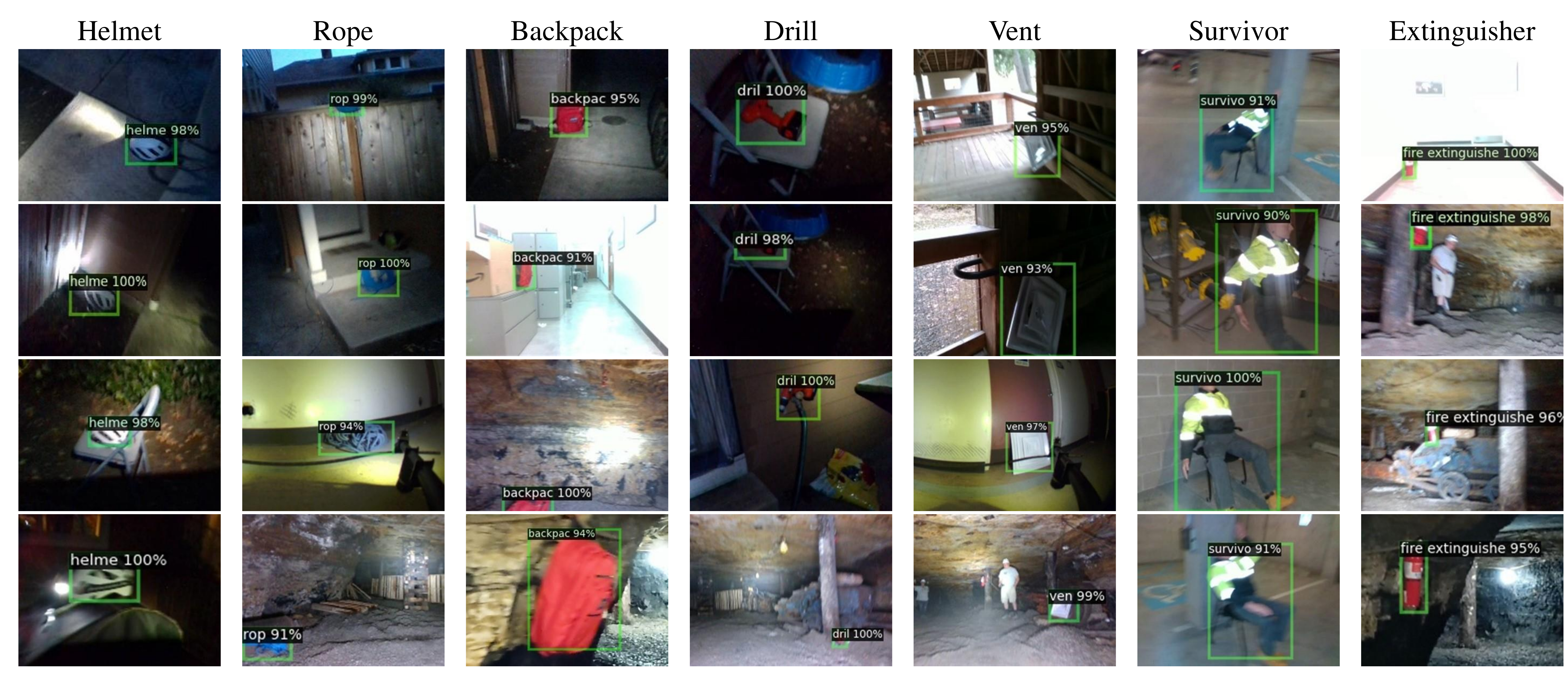}
	\caption{Representative examples of 3-shot detection of AirDet on the DARPA Subterranean Challenge. Provided with merely three support images for these unseen novel objects, AirDet can directly detect them with scale variation and partial occlusion in distinct environments and illumination conditions.}
	\label{fig:subt-app}
\end{figure*}

More qualitative detection results from the VOC-2012 \cite{everingham2010pascal} validation dataset and COCO \cite{lin2014microsoft} validation dataset are shown in \fref{fig:voco}. Provided with merely 3-shot support images per novel class, AirDet can directly detect unseen objects in various scales and distinct viewpoints from different environments.

We also exhibit more representative 3-shot detection results from the DARPA Subterranean Challenge \cite{subtchallenge} without fine-tuning in \fref{fig:subt-app}. For each novel class, \textit{i.e.}, helmet, rope, backpack, drill, vent, survivor, and fire-extinguisher, we have selected the objects from distinct environments including in-door, out-door, cave, tunnel, \textit{etc.} We find AirDet can maintain robust when faced with the challenging factors during exploration, \textit{e.g.}, illumination variation (examples from the helmet), partial occlusion (the second and third-row in the backpack), scale variation (examples from drill), and blur (the last row in survivor). Moreover, AirDet generally outputs high classification scores (higher than 0.9) and precise bounding boxes for the novel unseen classes, which demonstrate the promising prospect of AirDet for autonomous exploration tasks.

\section{More Deep Visualization}\label{sec:more_deep}

\begin{figure*}[!t]
	\centering
	\includegraphics[width=1\textwidth]{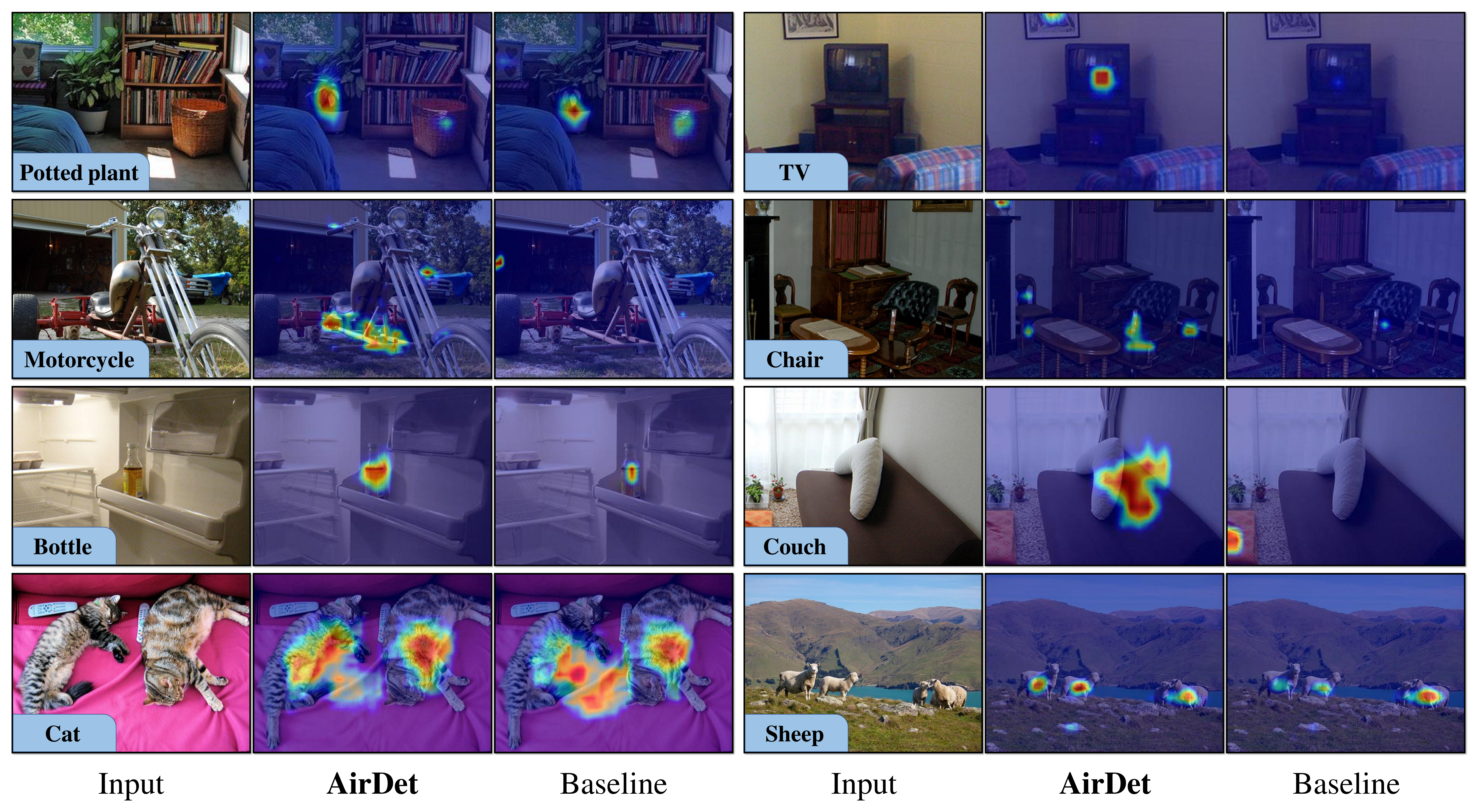}
	\caption{Deep visualization from the proposal generation module of AirDet and baseline \cite{fan2020few}. Compared with baseline, AirDet can better notice and concentrate on and novel object region, which leads to its more effective region proposals.}
	\label{fig:de_rpn}
\end{figure*}

\begin{table}[!t]
	\centering
	\setlength{\tabcolsep}{0.4mm}
	\fontsize{6}{7.5}\selectfont
	\caption{Detailed information about the 4 splits in LVIS dataset.}
	\begin{tabular}{cccc|cccc}
		\toprule
		\multicolumn{4}{c|}{Split1}   & \multicolumn{4}{c}{Split2} \\
		\midrule
		Class & Instance & Class &  Instance & Class &  Instance & Class & Instance \\
		bath mat & 63    & mousepad & 66    & ashtray & 51    & billboard & 270 \\
		birthday cake & 74    & pan   & 242   & taxi  & 68    & dresser & 39 \\
		blender & 57    & paper plate & 170   & duck  & 134   & figurine & 168 \\
		blouse & 99    & printer & 59    & guitar & 52    & hair dryer & 32 \\
		chandelier & 66    & saddle blanket & 94    & fume hood & 33    & polar bear & 36 \\
		Christmas tree & 72    & saucer & 103   & ottoman & 48    & pajamas & 54 \\
		grill & 90    & stool & 126   & radiator & 41    & scale & 46 \\
		mattress & 74    & tinfoil & 210   & shoulder bag & 61    & urinal & 237 \\
		\midrule
		\midrule
		\multicolumn{4}{c|}{Split3}   & \multicolumn{4}{c}{Split4} \\
		\midrule
		Class &  Instance & Class &  Instance & Class &  Instance & Class &  Instance \\
		blackboard & 37    & bridal gown & 23    & bear  & 116   & cistern & 182 \\
		bullet train & 25    & doormat & 28    & paper towel & 171   & parking meter & 282 \\
		fire engine & 42    & fish  & 92    & pickup truck & 209   & pot   & 121 \\
		hairbrush & 28    & kettle & 31    & saddle & 320   & saltshaker & 105 \\
		map   & 36    & piano & 24    & ski parka & 428   & soccer ball & 258 \\
		radio & 23    & teapot & 38    & statue & 204   & sweatshirt & 427 \\
		tongs & 44    & cover & 49    & tarp  & 160   & towel & 762 \\
		tripod & 24    & wallet & 29    & vest  & 168   & wine bottle & 223 \\
		\bottomrule
	\end{tabular}%
	\label{tab:lvis}%
\end{table}%

\begin{figure*}[!t]
	\centering
	\includegraphics[width=1\textwidth]{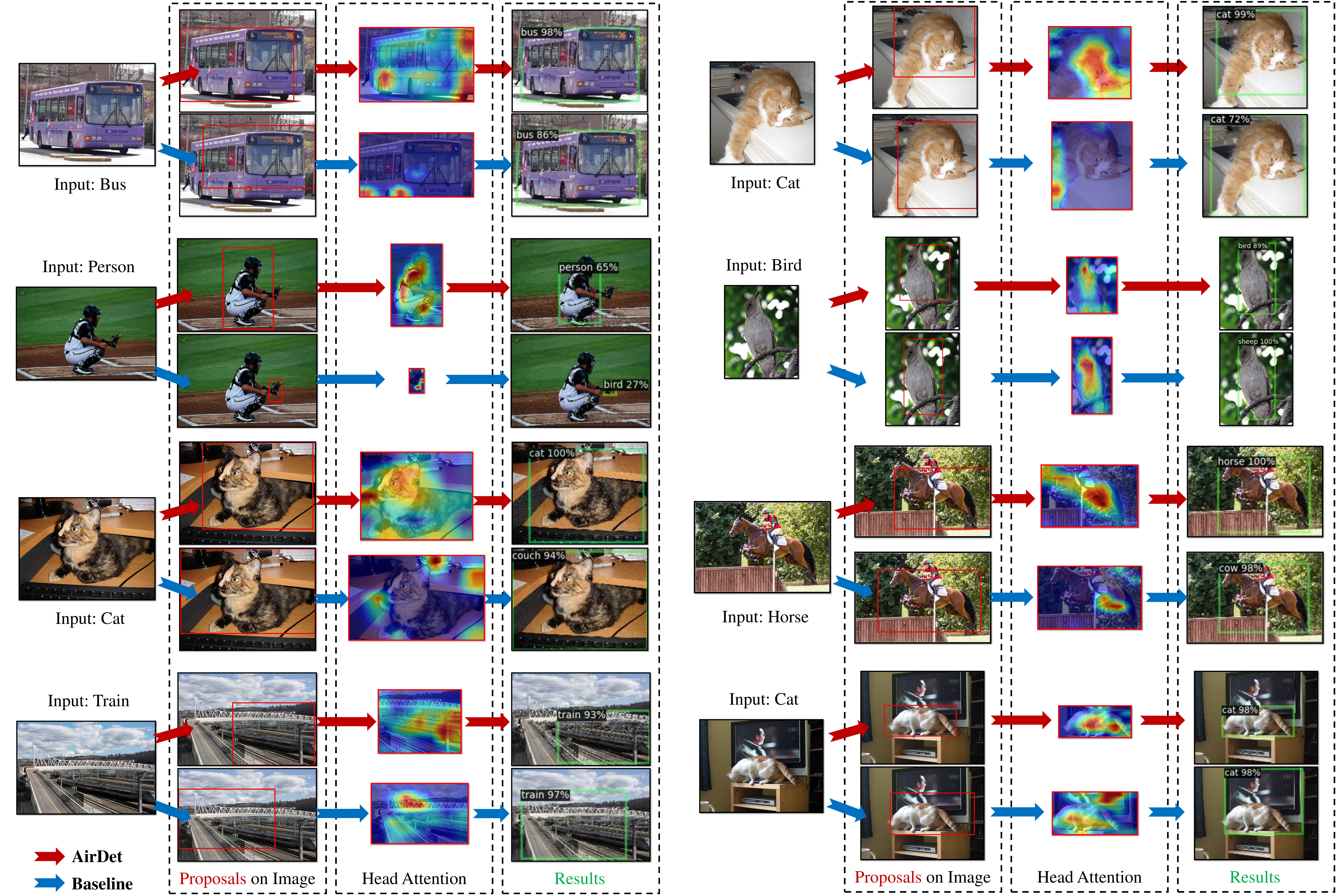}
	\caption{Deep visualization from the detection head of AirDet and baseline \cite{fan2020few}. With similar proposals in \red{red} boxes, AirDet can focus more precisely on the most representative part of the object, resulting in more accurate box regression and classification.}
	\label{fig:de_head}
\end{figure*}

To better demonstrate why the proposed AirDet works well, we present more deep visualization \cite{gradcam} with 5-shot supports for the proposal generation and the detection head.
As shown in \fref{fig:de_rpn}, we backpropagate the gradient of highest objectiveness score generated by AirDet and our baseline \cite{fan2020few} to the whole image. Compared with baseline, AirDet can better notice and concentrate on the novel region. For example, in the category ``cat", AirDet is not distracted by the carpet while our baseline loses its attention. Also, in ``sheep", AirDet can notice all the novel instances well but baseline may miss several instances. Therefore, by virtue of the SCS module, AirDet can generated more effective proposals.

In \fref{fig:de_head}, we backpropagate the gradient of highest classification score to the corresponding proposal image patch (\red{red} box). Thanks to the more representative class prototype from GLR and the fully relation-based detection head, AirDet is more capable of precisely predicting the category and box of an instance. For example, let's consider the cat at the bottom left. With similar proposal box, AirDet can concentrate better on the object region while baseline is distracted by its context. Consequently, the baseline incorrectly classifies a `cat' as a `couch'.
Another example is the cat at bottom right, since AirDet can focus better on the object, it produces a more accurate bounding box than baseline.

\section{Details About LVIS dataset}\label{sec:lvis}
We introduce the detailed information for LVIS \cite{gupta2019lvis} dataset. The class names and the number of instances for each class are shown in \tref{tab:lvis}.
It includes 64 classes and is sampled into 4 splits, each of which contains 890, 502, 365, and 1586 images, respectively.

\section{Detailed Limitations}\label{sec:de_limi}

\begin{table*}[!t]
	\centering
	\setlength{\tabcolsep}{2mm}
	\fontsize{6}{8.5}\selectfont
	\caption{Cross-domain performance on VOC-2012 validation dataset. AirDet is pre-trained with different number of classes (left) and instances (right).}
	\begin{tabular}{c|ccc|ccc|ccc}
		\toprule[1.2pt]
		Cls/Inst. & \multicolumn{3}{c|}{50/123,258} & \multicolumn{3}{c|}{55/140,682} & \multicolumn{3}{c}{60/148,872} \\
		\midrule
		Shots & AP    & AP$_{50}$  & AP$_{75}$  & AP    & AP$_{50}$  & AP$_{75}$  & AP    & AP$_{50}$  & AP$_{75}$ \\
		3     & 6.54  & 13.13 & 5.64  & 11.20 & 20.74 & 11.02 & \textbf{16.89} & \textbf{28.61} & \textbf{17.36} \\
		5     & 7.05  & 14.26 & 5.85  & 11.76 & 21.42 & 11.53 & \textbf{17.83} & \textbf{29.78} & \textbf{18.38} \\
		\bottomrule[1.2pt]
	\end{tabular}%
	\label{tab:diff_cls}%
\end{table*}%

\begin{table*}[!t]
	\centering
	\setlength{\tabcolsep}{2mm}
	\fontsize{6}{8.5}\selectfont
	\caption{Comparison of average results in the real-world tests using the same (left) or different (right) objects as support examples.}
	\begin{tabular}{c|cc|cc|cc|cc}
		\toprule[1.2pt]
		Metric & \multicolumn{2}{c|}{AP} & \multicolumn{2}{c|}{AP$_{50}$} & \multicolumn{2}{c|}{AR$_1$} & \multicolumn{2}{c}{AR$_{10}$} \\
		\midrule
		Support & same  & diff. & same  & diff. & same  & diff. & same  & diff. \\
		\textbf{AirDet} & \textbf{16.4} & \textbf{10.4} & \textbf{42.3} & \textbf{25.2} & \textbf{23.6} & \textbf{18.0} & \textbf{28.6} & \textbf{23.2} \\
		A-RPN & 12.5  & 9.4   & 31.9  & 21.7  & 20.3  & 15.6  & 23.8  & 21.4 \\
		\bottomrule[1.2pt]
	\end{tabular}%
	\label{tab:diff_support}%
\end{table*}%

\begin{table*}[!t]
	\centering
	\setlength{\tabcolsep}{0.6mm}
	\fontsize{6}{8.5}\selectfont
	\caption{3-shot per class AP results of AirDet without fine-tuning in COCO validation dataset and VOC-2012 validation dataset.}
	\begin{threeparttable}
    \begin{tabular}{ccc|ccc|ccc|ccc}
    \toprule[1pt]
    Category & COCO  & VOC   & Category & COCO  & VOC   & Category & COCO  & VOC   & Category & COCO  & VOC \\
    \midrule
    aeroplane & 11.5  & 17.7  & bicycle & 1.11  & 7.5   & pottedplant & 0.3   & 2.8   & sheep & 5.2   & 17.1 \\
    boat  & 2     & 1.0   & bottle & 4.1   & 12.1  & train & 3.5   & 11.5  & tvmonitor & 20.7  & 21.7 \\
    car   & 13    & 36.4  & cat   & 15.7  & 24.7  & bird  & 4.3   & 21.2  & dog   & 8.2   & 19.0 \\
    cow   & 2.4   & 14.0  & diningtable & 0.4   & 1.2   & bus   & 30.4  & 23.0  & person & 1.3   & 2.2 \\
    horse & 11.7  & 20.2  & motorbike & 5.4   & 4.1   & chair & 2.5   & 7.4   & sofa  & 9     & 14.1 \\
    \bottomrule[1pt]
	\end{tabular}\label{tab:perclass}%
	\end{threeparttable}
\end{table*}%
\myparagraph{Dependence of exhaustive base training:}
One potential limitation of AirDet is that it requires a large number of base classes during training to generalize. To exhaustively study this, we present the results on VOC-2012 validation dataset, which are obtained using models trained with different numbers of base classes. As shown in \tref{tab:diff_cls}, pre-training with fewer base classes and instances can make the model degrade.

\myparagraph{Dependence of high quality support images:} The robots in the real-world will utilize the \textit{online} defined objects (supports) to find the specific objects, where the supports are in good quality and appearance. Without this beneficial condition, the performance of AirDet will drop as shown in \tref{tab:diff_support}. Yet AirDet can still identify the novel objects in the tests compared with A-RPN \cite{fan2020few} even with significant appearance change.

\myparagraph{Result variance among different classes:} 
As aforementioned, the failure cases of AirDet in COCO and VOC datasets are mainly due to false classification, which results in high variance among different classes. We present the average precision for each novel class with a 3-shot evaluation setting in \tref{tab:perclass}. We observe that for novel classes like TV, monitor, and bus, the average precision can be up to 20 and 30. Nevertheless, for some other novel classes, \textit{e.g.}, boat, potted plant, and dining table, the scores are much lower. Such a high variance among different classes indicates the limitation of the classifier in the detection head. We observe that such a limitation also exists in other SOTAs \cite{fan2020few,xiao2020few,wu2020multi,wang2020frustratingly}, which guides our future work.
\end{document}